\documentclass[letterpaper, 10 pt, conference, twoside]{IEEEtran}
\IEEEoverridecommandlockouts
\usepackage{subcaption} 
\usepackage{cite}
\usepackage{amsmath,amssymb,amsfonts}
\usepackage{algorithmic}
\usepackage{graphicx}
\usepackage{textcomp}
\usepackage{xcolor}
\usepackage{adjustbox}
\usepackage{booktabs}
\usepackage{caption}
\usepackage{placeins}
\usepackage[colorlinks=false,
            pdfborder={0 0 1},
            linkbordercolor={1 0 0},   
            citebordercolor={0 1 0}    
           ]{hyperref}
\usepackage{cleveref}
\def\BibTeX{{\rm B\kern-.05em{\sc i\kern-.025em b}\kern-.08em
    T\kern-.1667em\lower.7ex\hbox{E}\kern-.125emX}}
\begin{document}

\captionsetup{skip=1pt}
\setlength{\textfloatsep}{1pt}
\setlength{\belowdisplayskip}{1pt} \setlength{\belowdisplayshortskip}{1pt}
\setlength{\abovedisplayskip}{1pt} \setlength{\abovedisplayshortskip}{1pt}
\setlength{\floatsep}{1pt} \setlength{\textfloatsep}{1pt}
\setlength{\intextsep}{1pt}
\setlength{\abovecaptionskip}{1pt}
\setlength{\belowcaptionskip}{1pt}
\captionsetup{belowskip=1pt}

\title{\LARGE \bf GraphSeg: Segmented 3D Representations\\via Graph Edge Addition and Contraction
}

\author{Haozhan Tang$^{1}$ \and Tianyi Zhang$^{1}$ \and Oliver Kroemer$^{1}$ \and Matthew Johnson-Roberson$^{1}$ \and Weiming Zhi$^{1,*}$
\thanks{$^{*}$email: {\tt\small wzhi@andrew.cmu.edu}.}%
\thanks{$^{1}$ Robotics Institute, Carnegie Mellon University, Pittsburgh, PA, USA}%
}

\maketitle

\begin{abstract}
Robots operating in unstructured environments often require accurate and consistent object-level representations. This typically requires segmenting individual objects from the robot's surroundings. While recent large models such as Segment Anything (SAM) offer strong performance in 2D image segmentation. These advances do not translate directly to performance in the physical 3D world, where they often over-segment objects and fail to produce consistent mask correspondences across views. In this paper, we present \emph{GraphSeg}, a framework for generating consistent 3D object segmentations from a sparse set of 2D images of the environment without any depth information. GraphSeg adds edges to graphs and constructs dual correspondence graphs: one from 2D pixel-level similarities and one from inferred 3D structure. We formulate segmentation as a problem of edge addition, then subsequent graph contraction, which merges multiple 2D masks into unified object-level segmentations. We can then leverage \emph{3D foundation models} to produce segmented 3D representations. GraphSeg achieves robust segmentation with significantly fewer images and greater accuracy than prior methods. We demonstrate state-of-the-art performance on tabletop scenes and show that GraphSeg enables improved performance on downstream robotic manipulation tasks. Code available at \href{https://github.com/tomtang502/graphseg.git}{https://github.com/tomtang502/graphseg.git}.
\end{abstract}

\section{Introduction}
Robots operating in diverse environments typically need to construct internal representations of their surroundings from their sensor inputs. In many cases, individual objects within these representations need to be segmented out to facilitate downstream interaction with these objects. This is particularly the case in grasping \cite{wright2024vprism} and motion generation \cite{GeoFab_gloabL_opt}. This paper tackles the problem of obtaining object-level representations via the segmentation of tabletop scenes, common to robot manipulation, from a set of multi-view 2D images of the tabletop. 

Advances in computer vision have led to large pre-trained models, most notably \emph{Segment Anything} \cite{kirillov2023segment}, which can efficiently segment a 2D image into a set of masks. However, these models do not extend directly into the segmentation of 3D representations or over a set of multi-view images. Pre-trained models that segment 2D images into semantically meaningful parts often \emph{over-segment}, that is, they may break down a consistent object into multiple masks. Additionally, it is challenging to accurately find correspondences between the masks over different images, and obtain masks which correspond to masks over the robot's 3D representation. 

In this paper, we propose \emph{GraphSeg}, a framework that produces segmented object-level representations. GraphSeg models the correspondence of masks over multi-view images as graphs, and then formulates the matching of segmentation masks as a graph edge addition and contraction problem. A graph contraction problem seeks to simplify a graph by removing edges and merging adjacent vertices to obtain super-vertices, which contain several of the original vertices. 

\begin{figure}[t]
\centering
    \includegraphics[width=0.45\textwidth]{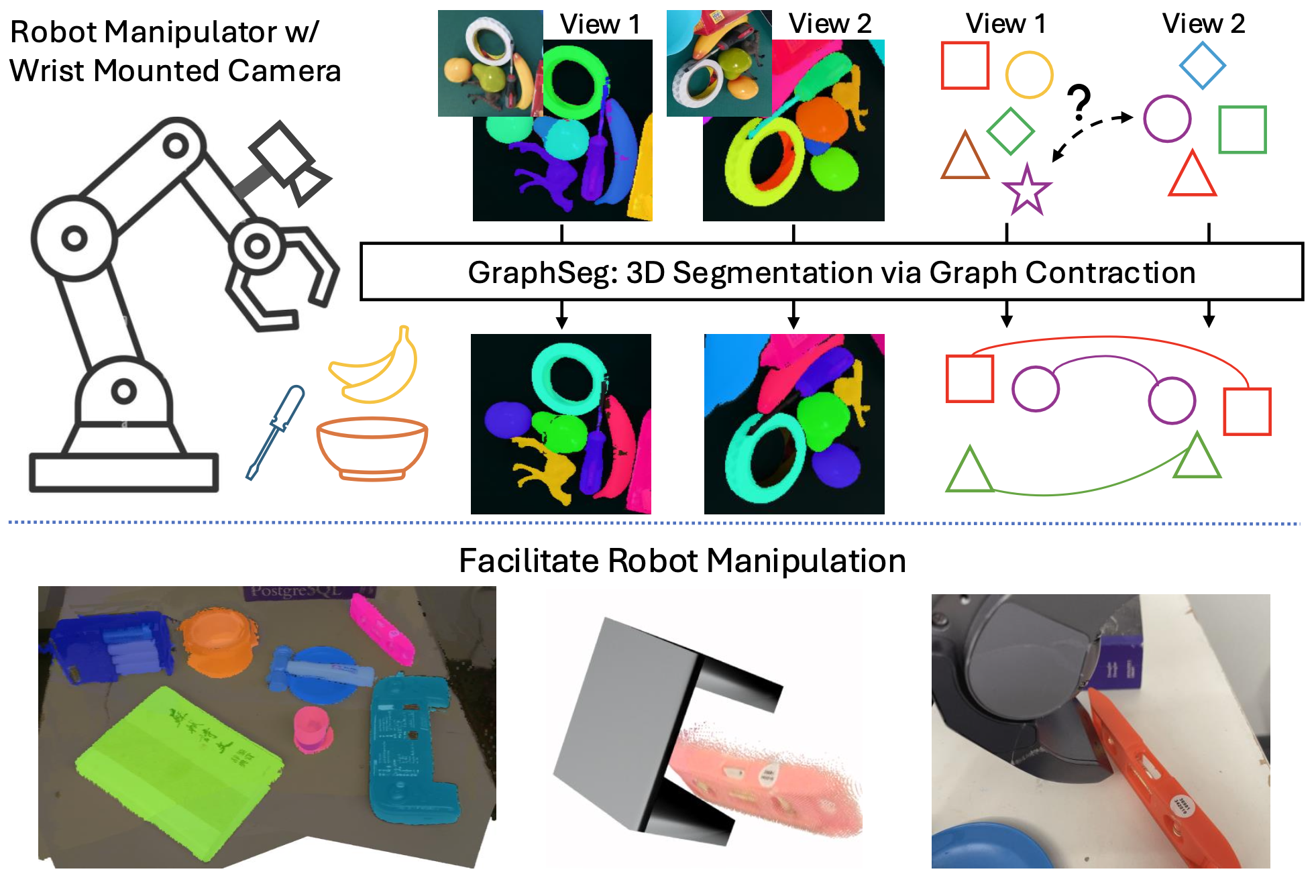}
    \caption{Robots often need to work with object-level representations. In this work, we tackle the problem of building segmented 3D scenes from a set of images, and introduce the GraphSeg framework. GraphSeg solves multi-view 3D segmentation via a novel graph edge addition and contraction procedure. This facilitates downstream robot manipulation}
    \label{fig:teaser}
\end{figure}

GraphSeg adds edges by considering the correspondence masks, based on both pixel-level correspondences over the 2D images, as well as inferred correspondences over the underlying 3D structure of the environment. We demonstrate that by formulating and efficiently solving our graph edge addition and contraction problem, GraphSeg can reliably produce consistent 3D segmentations, is robust to over-segmentation, and can operate even with very few images (sparse-view), outperforming existing methods by a large margin. Additionally, the fully segmented structure is produced by leveraging \emph{3D foundation models}, which learn to recover 3D representations from multi-view images, and no depth readings from sensors are required. Concretely, our technical contributions are as follows:
\begin{enumerate}
\item The GraphSeg framework capable of producing cleanly segmented 3D scene representations;
\item The formulation of 3D segmentation as a problem of graph edge addition and contraction, along with algorithms to efficiently solve the graph contraction;
\item Extensive empirical evaluation showing state-of-the-art 3D segmentation performance, and demonstrating utility to downstream robot manipulation tasks.
\end{enumerate}

\begin{figure*}[t]
\centering
    \includegraphics[width=0.95\textwidth]{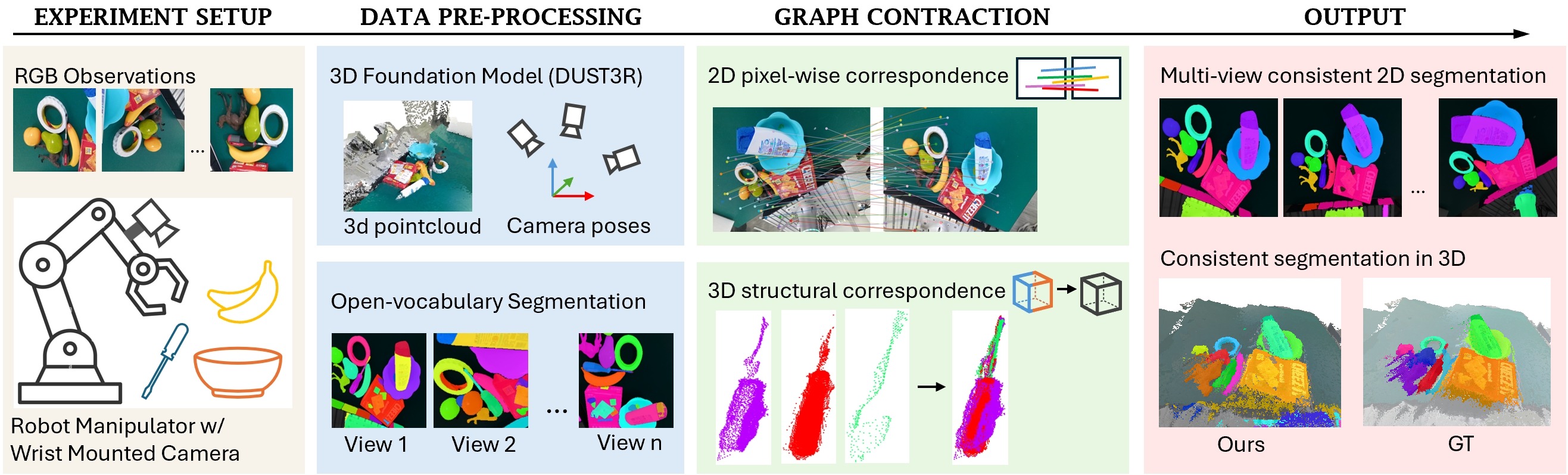}
    \caption{GraphSeg enables consistent 3D segmentation. We can obtain a set of segmented 2D images by leveraging the pre-trained open-vocabulary segmentation model. We then leverage edge addition via correspondence and graph contraction, over both 2D and lifted 3D representations, to obtain segmented 3D representations.}
    \label{fig:overview}
    \vspace{-1em}
\end{figure*}

\section{Related Work}
Early deep learning methods for image segmentation were driven by advances in Convolutional Neural Networks, including Deeplab~\cite{chen2017deeplab} and SegNet~\cite{badrinarayanan2017segnet}. With the emergence of 3D datasets~\cite{dai2017scannet, fang2020graspnet} and the increasing demand for robot manipulation, many approaches have been proposed for 3D instance segmentation to support control and planning. These methods often fall into three categories: (1) closed-set 3D instance segmentation, (2) deep learning-based open-vocabulary 3D instance segmentation, and (3) zero-shot open-vocabulary 3D instance segmentation.

\textbf{Close-set 3D instance segmentation} methods operate on a fixed set of classes, assigning parts of the 3D representation to each class~\cite{vu2022softgroup, hu2021bidirectional, schult2023mask3d, hu2021vmnet, han2020occuseg}. These models use prior knowledge of predefined classes and perform poorly in open-vocabulary scenarios. Our method is designed for open-vocabulary settings without predefined instance priors.

\textbf{Deep learning-based open-vocabulary 3D instance segmentation} approaches train on annotated 3D segmentation datasets to predict instance masks rather than assigning regions to fixed classes~\cite{zhou2024point, takmaz2023openmask3d}. However, due to limited 3D instance data, these models often fail to generalize in real-world tabletop scenarios without fine-tuning. Our method aims to generalize zero-shot to arbitrary tabletop settings for fixed-base robot manipulators. Although 3D data remains limited, the computer vision community has produced a wealth of 2D segmentation datasets~\cite{cordts2016cityscapes, everingham2010pascal, kirillov2019panoptic, caesar2018coco, zhou2017scene, arbelaez2010contour, silberman2012indoor}, enabling robust class-agnostic 2D segmentation via foundation models like CropFormer~\cite{qilu2023high} and SAM~\cite{ravi2024sam}.

\textbf{Zero-shot open-vocabulary 3D instance segmentation} methods such as MaskClustering~\cite{yan2024maskclustering}, SAM3D~\cite{yang2023sam3d}, and OVIR-3D~\cite{lu2023ovir} leverage 2D segmentation models and aggregate segmented point clouds using 3D spatial relationships. These are most comparable to our proposed GraphSeg, which we evaluate against. Other approaches~\cite{zhu2023pointclip, nguyen2024open3dis} project 3D data to 2D and apply visual-text embeddings like CLIP~\cite{radford2021learning} and DINO~\cite{zhang2022dino} to guide segmentation. However, 3D-centric methods may fail in sparse view settings, and embedding-based methods struggle with semantically similar objects. Our approach differs by formulating segmentation as a graph edge addition and contraction problem, leveraging both optical 2D and structural 3D correspondence for improved reliability.

\section{Preliminaries}

GraphSeg makes use of advances in \emph{foundation models}. Foundation models are large deep learning models trained on large and diverse datasets, and are intended as plug-and-play modules to facilitate downstream operations in a zero-shot manner \cite{Bommasani2021FoundationModels}. In this work, we explore foundation models applied to segmenting 2D images, and constructing 3D representations. Here, we will briefly elaborate on pre-trained segmentation models, 3D foundation models, as well as \emph{graph contraction} which is embedded into our GraphSeg formulation.

\subsection{Pre-trained Segmentation Models}
Segment Anything (SAM)\cite{ravi2024sam} is an open-world segmentation model that produces high-quality mask proposals for any object in an image without relying on predefined semantic categories. Given an rgb image $I \in \mathbb{R}^{H \times W \times 3}$, SAM automatic mask generator output a set binary mask proposals
\begin{align}
    \text{SAM}(I) = \{ m_1, m_2, \ldots \}, m_i \in \{0,1\}^{H \times W} .
    \label{eq:sam}
\end{align} It is trained on a large and diverse 2D dataset so that it generalizes across domains in 2D single image segmentation without being restricted to any semantic class. Combining with DINO\cite{zhang2022dino} visual-semantic embedding, we can use a semantic prompt to reliably extract a specific instance mask, such as the bare tabletop from a single image.

\subsection{3D Foundation Models}
3D Foundation models projects sets of 2D images into 3D. Suppose we have a pair of RGB images with width $W$ and height $H$, i.e. $I_{1},I_{2} \in \mathbb{R}^{W\times H\times 3}$, 3D foundation models, such as DUST3R\cite{DUSt3R_cvpr24} specifically, produces pointmaps $X^{1,1}, X^{1,2} \in \mathbb{R}^{W \times H \times 3}$, which map each 2D pixel to its predicted 3D coordinates aligned under the first image's coordinate frame. It also generates confidence maps $C^{1,1}, C^{1,2} \in \mathbb{R}^{W \times H}$ for each pointmap to quantify the uncertainty in the foundation model’s predictions at each pixel. By matching each pixel’s predicted 3D coordinates in one pointmap with the nearest coordinates in the corresponding pointmap, dense correspondences between pixels in the image pair can be established without relying on handcrafted features. The point cloud of the pair of $I_{1}$ and $I_{2}$ are then aligned to reversely predict the corresponding camera poses $P_1, P_2$. As a result, we can obtain pixel-wise correspondence, an aligned 3D point cloud, camera poses by using these foundation models as a black box function. 3D foundation models have been used to understand the geometry of objects during grasping \cite{sim_grasp}, and can enable the construction of 3D photorealistic representations \cite{kerbl3Dgaussians,zhang2024darkgs,zhang2024recgs}.
\begin{figure}[t]
\centering
    \includegraphics[width=0.8\linewidth]{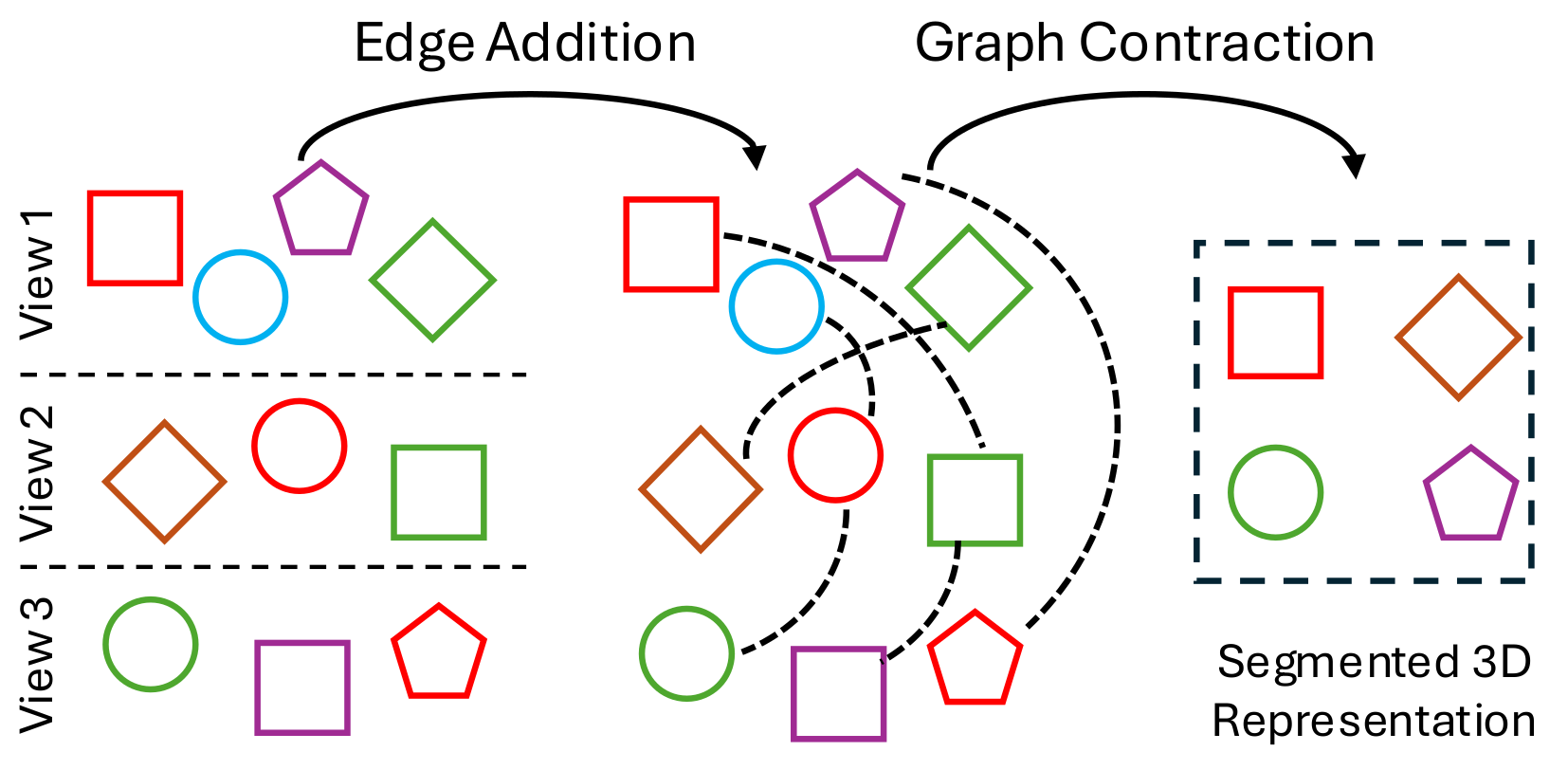}
    \caption{At the core of GraphSeg is an edge addition and graph contraction process. The edge addition is achieved by finding correspondences between masks, via pixel-to-pixel features and 3D structural information.}
    \label{fig:graph_edge_contr}
\end{figure}

\subsection{Graph Contraction}
In a typical graph, we have a set of vertices and edges. A graph partition divides the graph into subsets, where each subset consists of vertices that are associated with a representative vertex of that partition. The process of graph contraction—a classical concept in theoretical computer science—aims to produce such a partition by mapping together vertices that are connected, grouping them into the same subset. Formally, for a given graph $(V, E)$, \begin{align} \mathrm{GraphContract}(V, E) ;\mapsto; f, \quad \text{where } f \colon V ;\to; \tilde{V}, \end{align} and $\tilde{V}$ denotes the set of representative vertices corresponding to the graph partition.

\section{GraphSeg}
Here, we are assumed to have a fixed-based robot manipulator with a mounted camera operating in a classical tabletop setup. We control the end-effector manipulator to take a small set of $N$ RGB images of the tabletop, $\{I_{1},\ldots, I_{N}\}$ where $I_i \in \mathbb{R}^{H \times W \times 3}$. We leverage the large open-world segmentation model, Segment Anything \cite{kirillov2023segment}, to obtain 2D segmented masks from each image.

GraphSeg aims to output the corresponding segmented point cloud representation of the tabletop in the form of: (1) the relative aligned camera poses $\{P_{1},\ldots, P_{N}\}$; (2) 3D point clouds associated with each image, $\{\hat{X}_{1},\ldots, \hat{X}_{N}\}$; (3) Segmentation image $\{\hat{M}_{1},\ldots, \hat{M}_{N}\}$ that corresponds to the result of segmenting the point cloud, where each point is assigned a mask class that is consistent over all of the images. We can then extract individual objects in the scene by extracting the corresponding 3D points associated with the object class.

\begin{figure}[t]
    \centering
    \includegraphics[width=0.8\linewidth]{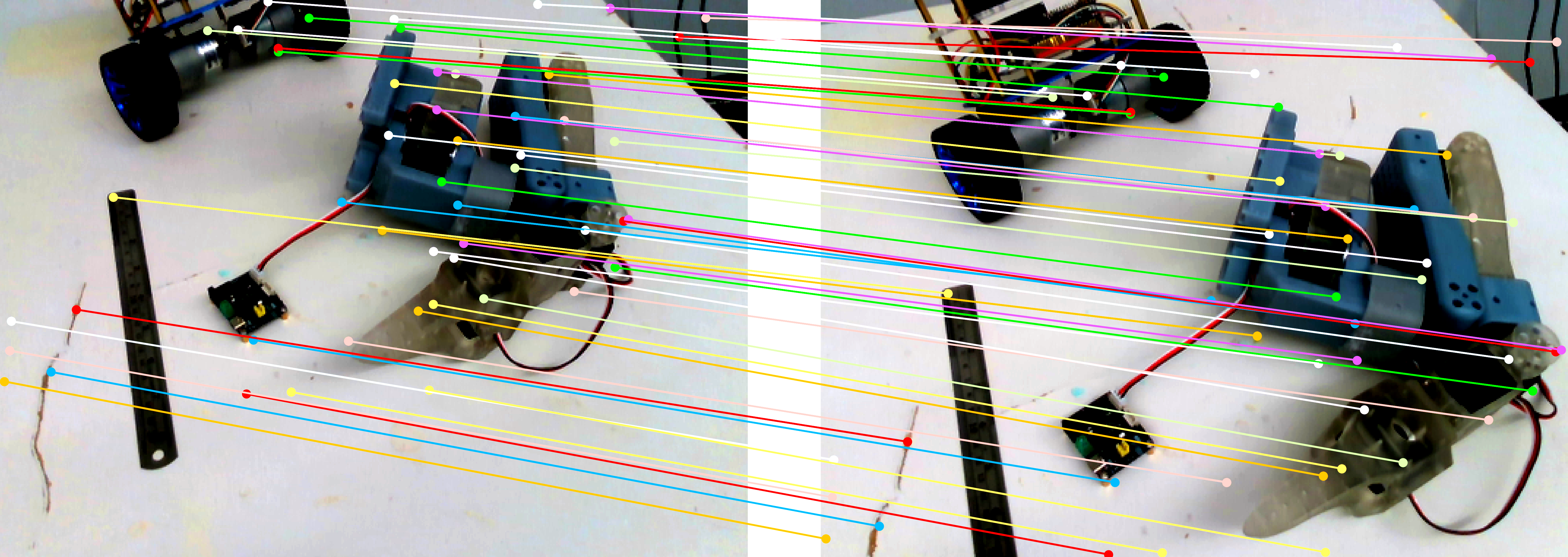}
    \caption{We can find correspondence between masks by considering pixel-level correspondence between images. Here we see examples of some correspondences between two images at different views.}
    \label{fig:pixel_cor}
\end{figure}

\subsection{Overview}
The overview of our method is given in \Cref{fig:overview}. The crux of our method lies in finding correspondences between the abundant segmentation mask classes over a set of multi-view images. The main challenges in extending 2D segmentation to consistent multi-view image segmentation are that each of the images may contain different sets of objects, and that over-segmentation may occur in many of the images. 

\begin{figure}[t]
\centering
    \includegraphics[width=0.9\linewidth]{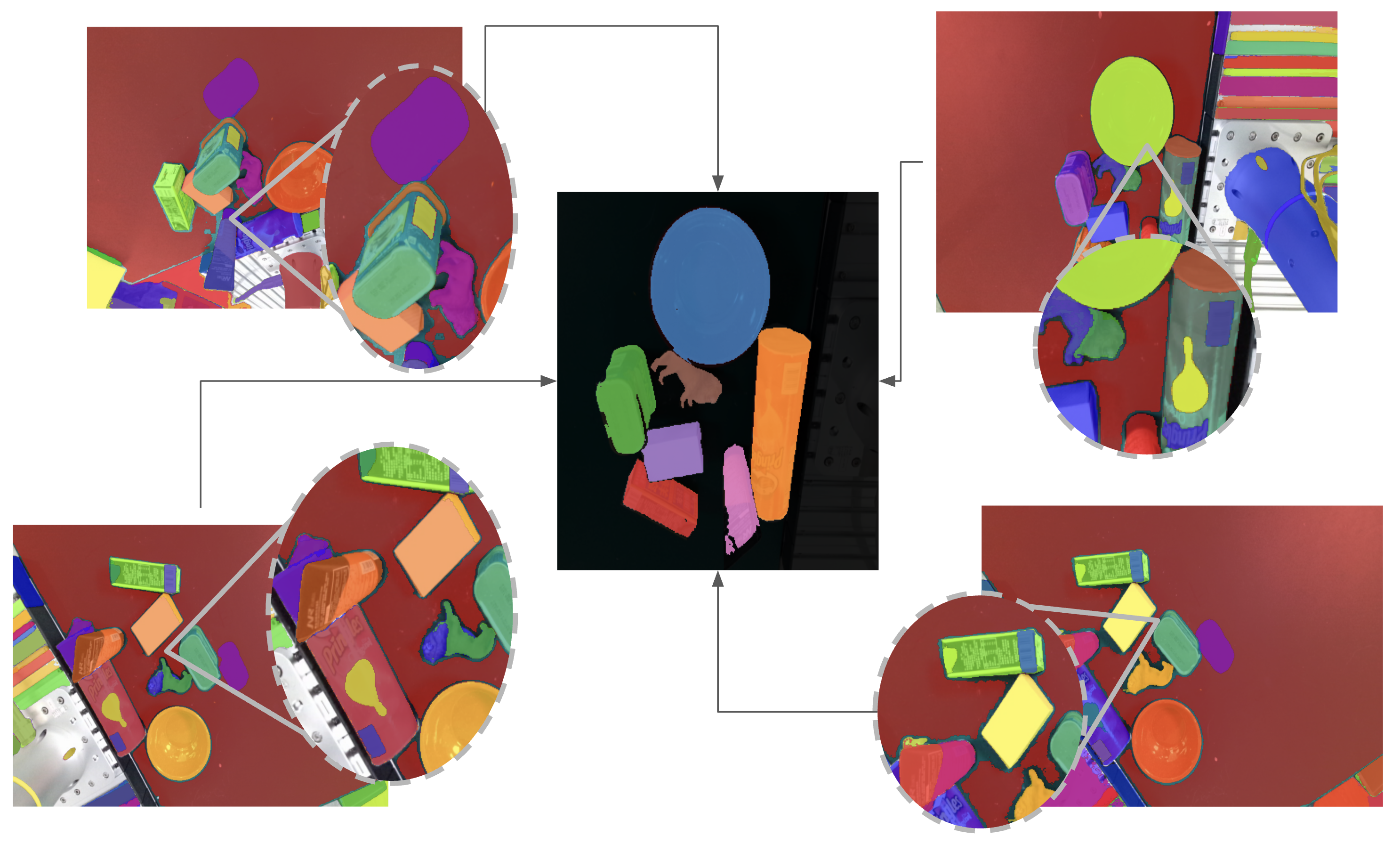}
    \caption{GraphSeg can take over-segmented 2d images (e.g. the labels have been unnecessarily segmented from the can and bottle) and produce consistent 3D segmentations (shown in the center).}
    \label{fig:robust_no_overseg}
\end{figure}

Our key insight is to formulate the matching of potentially inconsistent segmentation masks over multiple images as graph edge addition and contraction problems. We begin by assuming that each individual class in a mask of each image is a vertex. Through edge addition and graph contraction, we reduce the set of vertices, where each new vertex is a combination of previous vertices. The process of edge addition and graph contraction is illustrated in \cref{fig:graph_edge_contr}. Vertices are contracted together if they are masks of the same object captured over different views, and the final set of vertices corresponds to each segmented object, as shown in \cref{fig:robust_no_overseg}.

Here, we seek to leverage both the image-level information and the estimated 3D spatial properties. This is achieved by constructing dual correspondence graphs through an edge addition process. One of these graphs estimates 2D pixel-level correspondences, and the other correspondence graph is constructed based on the 3D structure captured in each image, obtained by lifting the 2D image representation into 3D via 3D foundation model. After the edge addition process, these graphs are \emph{contracted} with connected vertices being absorbed into a single vertex.

\subsection{Pixel-to-Pixel 2D-based Edge Addition}
We seek to leverage the visual information between the images in the collected image set to produce initial correspondences between masks from different images. Here, we each pass a pair of images into a pre-trained dense feature matching model \cite{edstedt2024roma}. We obtain a set of pixel-pairs across the two different input images, along with confidence estimates. We keep all confident pairs of pixel matches, and denote the set of matching pixels by $\mathcal{F}$. \Cref{fig:pixel_cor} illustrates several pixel-level matches between two images captured as different camera views. We add edges between two masks if the proportion of matching pixels in the mask exceeds a proportion threshold. That is, let $h(m_{1},m_{2})$ be the number of matching pixel pairs detected between masks $m_{1}$, $m_{2}$, i.e. 
\begin{align}
h(m_{1},m_{2}) &= \lvert\{(p, p'), (p, p')\in \mathcal{F}\lvert p\in m_{1}, v\in m_{2} \}\lvert.
\end{align}
We add the edge between the vertices $m_{1}$ and $m_{2}$, if the number of matched pixels relative to the size of the smaller mask exceeds a threshold, i.e., 
\begin{align}
\frac{h(m_{1}, m_{2})}{\min(g(m_{1}), g(m_{2}))}\geq\tau_{2D},
\end{align}
where $g(\cdot)$ is the pixel area of the mask, and $\tau_{2D}$ is a threshold for which we consider two masks to be correspondent and assign an edge. We then have a graph $G_{2d}=(V, E)$, where $V$ contains the set of all 2D segmentation masks, and $E$ contains all of the correspondence edges found by pixel-to-pixel matching. Here, we note that due to over-segmentation, a mask from one image may match with multiple masks in another image. The graph $G_{2d}$ is then contracted to obtain the set $\{m'_{1},\ldots m'_{p}\}$ where each $m'$ is a super-vertex contains multiple 2D masks, which have been identified to be the same object, over the image set and corresponds to an initial estimate of the 3D segmentation.

\subsection{Structural 3D-based Edge Addition}
Advances in learning-based dense 3D reconstruction have enabled even sparse sets of multi-view images to be \emph{lifted} into 3D. We seek to leverage the structural information of the 3D representations to further refine the initial segmentation obtained from contracting the pixel-to-pixel correspondence graph. Using only 2D pixel-to-pixel graph contraction, the resulting multi-image mask can segment the point cloud reasonably well. However, we can still observe over-segmentation in the set of super-vertices. A key reason is due to excessive surface detail. The initial 2D segmentation model often produces multiple small masks on the same object surface in many images. This typically occurs for objects with rich surface features, making it difficult for the generator to determine whether surface regions still belong to the same object across views.

We add further graph edges over the set of super-vertices. As the 3D foundation model produces a 3D point for each pixel in the input images, for each super-vertex, there is a one-to-one mapping to a point cloud. We measure the directed Chamfer distance of the corresponding points in 3D between two super-vertices, given by 
\begin{align}
D(X, Y) = \frac{1}{|X|}\sum_{\mathbf{x}\in X}\min_{\mathbf{y}\in Y}\lvert\lvert \mathbf{x}-\mathbf{y}\lvert\lvert_{2}^{2},
\end{align}
where $X$ and $Y$ are 3D point clouds corresponding to the super-vertex masks. A small distance indicates that $X$ is subsumed in $Y$ and we can connect an edge correspondence, giving us a resulting graph $G_{3D}=(\hat{V},\hat{E})$. Here, $\hat{V}=\{m'_{1},\ldots m'_{p}\}$ contains the solutions from contracting the pixel-to-pixel correspondence graph, and $\hat{V}$ contains edges if $D(m'_{1}, m'_{2})\leq\tau_{3D}$, for any $m'_{1}\neq m'_{2}$. Running graph contraction on the corresponding $G_{3D}$ produces our solution super-vertices which each represents a 3D segmentation. 

\subsection{Graph Contraction via Randomized Star}
To contract the constructed correspondence graphs, we leverage a variant of \textbf{randomized star contraction algorithm}, designed for efficient graph reduction. Each vertex $v$ in the graph  $G = (V, E)$ is assigned a random label $L(v) \in [0,1]$ to break symmetry. For every vertex, define its closed neighborhood as
\begin{align}
N[v] &= \{v\} \cup \{ u \mid (u,v) \in E \}.
\end{align}
A vertex becomes a star center if its label is the smallest in its neighborhood:
\begin{align}
L(v) &= \min_{u \in N[v]} L(u).
\end{align}
Otherwise, it contracts to the neighbor with minimum label:
\begin{align}
v^* &= \arg\min_{u \in N[v]} L(u),
\end{align}
and the edge $(v, v^*)$ is contracted, merging $v$ into the supervertex represented by $v'$. This contraction step is performed in parallel across all vertices and recursively applied to the resulting graph until a trivial structure remains. The use of random labels $L$ ensures unbiased decision-making and balanced contraction, preserving essential connectivity while significantly reducing the graph's complexity.

After obtaining our super-vertices, we can easily obtain the entire 3D point cloud of any segmented object simply by specifying the 2D mask of the object in the set of images, by searching through each super-vertex for the ID of the 2D mask and checking which super-vertex the mask was subsumed into.

\subsection{Efficient Implementation Heuristics}
When implementing GraphSeg, we can leverage several heuristics to enable more efficient and robust performance. We outline several implementation details in GraphSeg.

\textbf{Background Class:} We can leverage the properties of the tabletop manipulation problem structure for greater efficiency. We introduce a background class, which is captured by a single mask. As the 3D foundation model provides pixel-wise depth estimate, pixels that are significantly far out of the manipulation's reach will be filtered out and along with the table surface to be added into the background class.

\textbf{Speeding up the search for pixel-to-pixel correspondences:} We can build highly connected graphs without resorting to iterating over all the masks in other images. Given the set of input images, for each image we only find pixel-wise correspondence with images taken at a ''nearby'' camera pose. As the 3D foundation models provide both 3D structure and the camera poses where the images were taken, when considering the correspondences of an image, we can filter out images which we taken from a vastly different pose, as it would be exceedingly unlikely that they contain pixel-level correspondences.

\begin{table*}[t]
  \centering
  \caption{Quantitative comparison: We exhaustively test the performance of our method and compare it to the baseline method on the Graspnet-1B-main, Graspnet-1B-similar, and Graspnet-1B-unseen datasets \cite{fang2020graspnet}.}
  \label{tab:results}
  \begin{tabular}{l *{4}{c} @{\hspace{1em}} *{4}{c} @{\hspace{1em}} *{4}{c}}
    \toprule
    & \multicolumn{4}{c}{Graspnet-1B-main} & \multicolumn{4}{c}{Graspnet-1B-similar} & \multicolumn{4}{c}{Graspnet-1B-unseen} \\
    \cmidrule(lr){2-5} \cmidrule(lr){6-9} \cmidrule(lr){10-13}
    Method & IOU & $F1$ & $d_{\text{chamfer}}$ & $\text{IoU}_{sel}$ & IOU & $F1$ & $d_{\text{chamfer}}$ & $\text{IoU}_{sel}$ & IOU & $F1$ & $d_{\text{chamfer}}$ & $\text{IoU}_{sel}$ \\
    \midrule
    SAM3D                          & 0.319  & 0.4484 & 0.0556 & 0.317  & 0.3721 & 0.4877 & 0.0539 & 0.3708  & 0.346  & 0.4814 & 0.0625 & 0.3444 \\
    MaskClustering                 & 0.0137 & 0.0311 & 0.0421 & 0.0125 & 0.0133 & 0.0309 & 0.0337 & 0.0138  & 0.0132 & 0.0315 & 0.0461 & 0.0109 \\
    $\text{GraphSeg}^-$ (ours)      & 0.3916 & 0.5904 & \textbf{0.0042} & 0.3549 & 0.3966 & 0.5814 & \textbf{0.0031} & 0.3656  & 0.373  & 0.5563 & \textbf{0.0027} & 0.348 \\
    GraphSeg (ours)                & \textbf{0.5945} & \textbf{0.7595} & 0.0142 & \textbf{0.5816} & \textbf{0.6522} & \textbf{0.7997} & 0.0131 & \textbf{0.6418}  & \textbf{0.7308} & \textbf{0.848} & 0.0065 & \textbf{0.7281} \\
    \bottomrule
  \end{tabular}
  \vspace{-1.5em}
\end{table*}
\begin{table}[t]
  \centering
  \begin{tabular}{lccc}
    \toprule
     Method & SAM3D & MaskClustering & GraphSeg (ours) \\ \midrule
    Graspnet-1B-main & 0.3714 & 0.3815 & \textbf{0.6982} \\ 
    Graspnet-1B-similar & 0.4146 & 0.3971 & \textbf{0.743} \\ 
    Graspnet-1B-unseen   & 0.3864 & 0.4025 & \textbf{0.8121} \\ 
    \bottomrule
  \end{tabular}
  \caption{Each of the different methods filters out low-confident pixels at different rates. Here, we evaluate the quality of segmentation based on precision across all scenes. This is not sensitive to the number of points filtered.}\label{tab:compare_precision}
\end{table}

\section{Empirical Evaluations}
In this section, we empirically evaluate the efficacy and robustness of GraphSeg. We both qualitatively compare GraphSeg against state-of-the-art 3D segmentation methods, and quantitatively illustrate the robustness of GraphSeg to over-segmentation by the 2D segmentation model. Additionally, we stress test GraphSeg in the sparse multi-view setting, where there is limited overlapping visual features between the images. Many comparison methods filter away pixels from the inputs about which the methods are unsure. Next, we investigate the \emph{pixel utility} of various methods, that is, how many pixels remain in the final result. Finally, we demonstrate the applicability of GraphSeg for grasping objects within a tabletop environment.

\subsection{Experimental Setup}
\textbf{Datasets and Implementation:}
GraspNet-1Billion \cite{fang2020graspnet} dataset is a large and comprehensive dataset which consists of three large subsets of objects (main, similar, and unseen) and 97,280 images, along with camera poses, and fine-grained instance annotations on each pixel, which make it an optimal choice to evaluate our 3D segmentation method. The dataset has been carefully labelled, with a ground truth class assigned to each pixel. GraphSeg uses vision information only, and does not require any depth information included in the data. All of our experiments were conducted on a computer with Intel i9-14900HX CPU and NVIDIA GeForce RTX 4090 GPU. We use the default parameters provided by the original authors for the comparison methods, MaskClustering and SAM3D; We select the following hyperparameters for GraphSeg: the threshold for edge addition based on 2D correspondence is selected as the value of the $78^{th}$ percentile of all considered values. During the pixel-to-pixel edge construction, we subsample the number of sampled 2D correspondence points to 10000. The threshold for adding correspondence in the 3D edge addition step is given by $\tau_{3D} = 5e-4$. In practice, we have observed that GraphSeg is not overly sensitive to hyper-parameters, with values in the rough ballpark providing strong results.

\textbf{Metrics:} 
To evaluate the quality of 3D segmentation, we first want to find the subset of the segmented point cloud that corresponds to the target object. For each object in a scene, we find the corresponding point cloud by selecting (1) the segmented object class with the highest intersection over union (IoU) or (2) finding the segmented object class with the lowest chamfer distance, where the chamfer distance is defined by 
\begin{align}
    d_{\text{chamfer}}(S,S')\!=\!\sum_{x \in S}\!\min_{y \in S'} \| x\!-\!y \|_2^2 
    \!+\!\sum_{y \in S'}\!\min_{x \in S} \| y\!-\!x \|_2^2 .
    \label{eq:chamfer_d}
\end{align}
We compute the IoU and $F1$ score between the ground truth object point cloud, and the corresponding segmented object point cloud found via highest IoU, denoted as IoU and $F1$. We also report the chamfer distance (denoted as $d_{\text{chamfer}}$) and IoU (denoted as $\text{IoU}_{sel}$) between the ground truth object point cloud and the corresponding segmented object point cloud found via the lowest chamfer distance. To show the necessity of considering correspondences based on both 2D structural 3D information, we also report the same metrics compute on the output of our method without the 3D graph edge addition and contraction, denoted as $\text{GraphSeg}^-$. The mean of IOU, $F1$, $d_{\text{chamfer}}$, and $\text{IoU}_{sel}$ over all objects in all scenes in each dataset is reported in \cref{tab:results}.

\subsection{Segmentation Quality}

\begin{figure}[t]
  \centering
  \begin{subfigure}[t]{0.16\textwidth}
    \centering
    \includegraphics[width=0.98\linewidth]{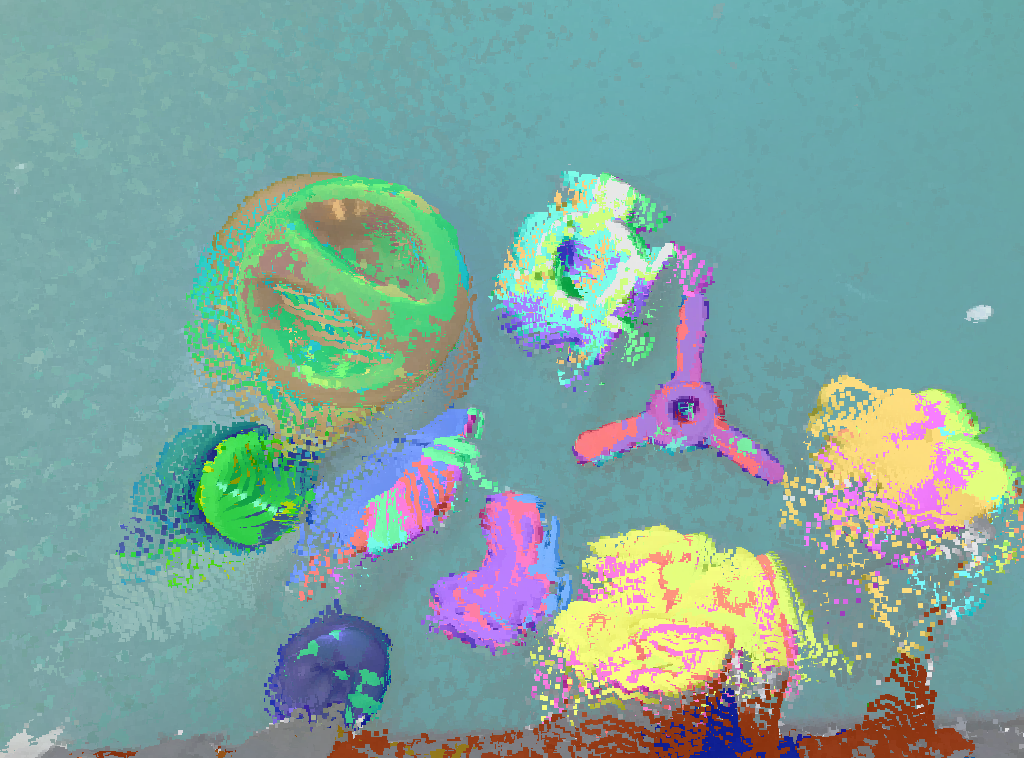}
    \caption{GraphSeg w/o 3D}
  \end{subfigure}%
  \begin{subfigure}[t]{0.16\textwidth}
    \centering
    \includegraphics[width=0.98\linewidth]{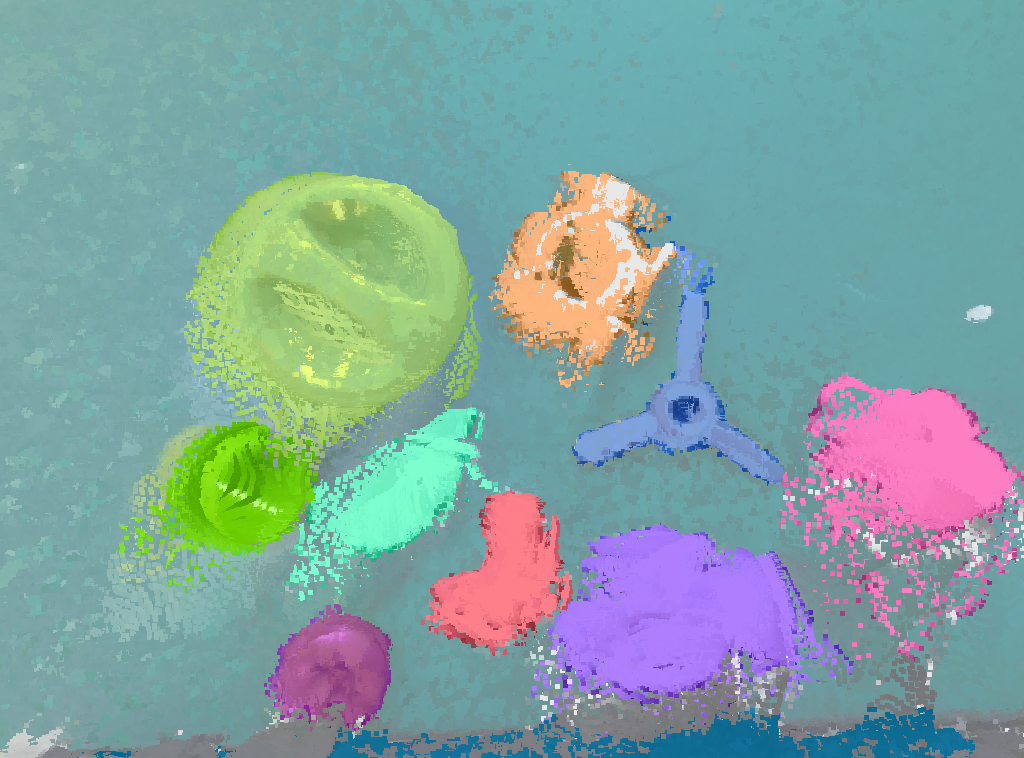}
    \caption{GraphSeg}
  \end{subfigure}%
  \begin{subfigure}[t]{0.16\textwidth}
    \centering
    \includegraphics[width=0.98\linewidth]{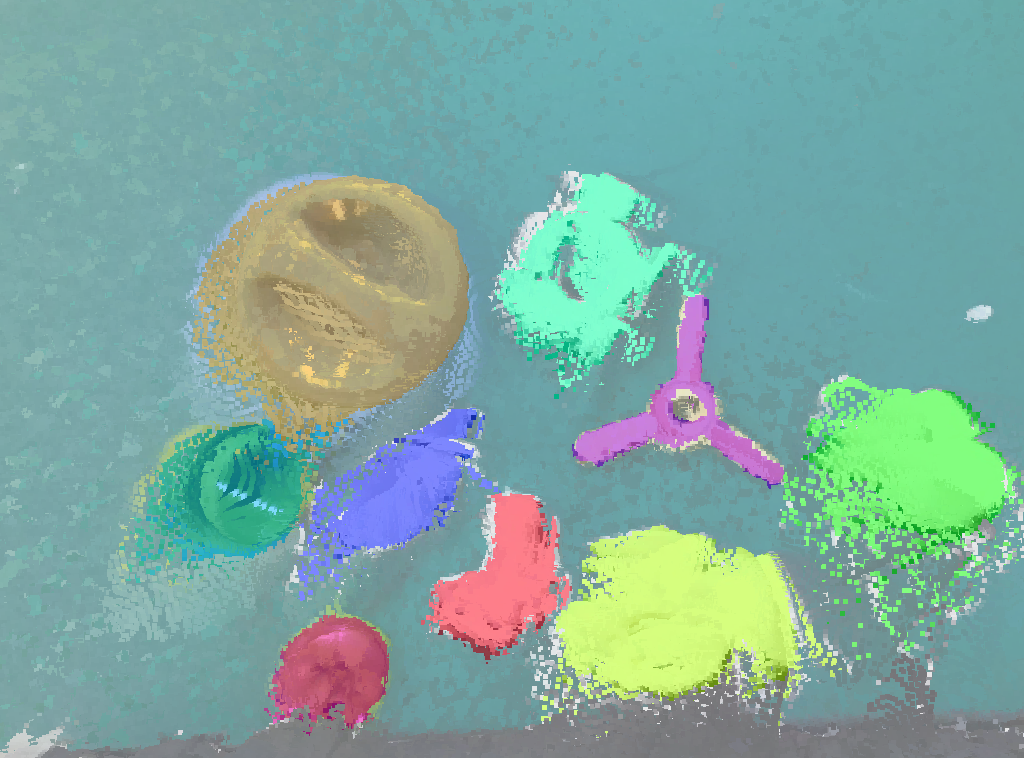}
    \caption{Ground Truth}
  \end{subfigure}%
  \caption{Although GraphSeg without 3D edge addition and graph contraction can identify individual objects, producing a small chamfer distance. It over-segments on feature-rich objects. This results in multiple masks being assigned to a single 3D object, as shown in (a). This is not the case after considering structural information as shown in (b).}
  \label{fig:ablation}
\end{figure}
As the quantitative result shown in \cref{tab:results}, our method achieved the state-of-the-art performance in all metrics except $d_{\text{chamfer}}$ in all datasets. Although the best $d_{\text{chamfer}}$ is reached by $\text{GraphSeg}^-$, it is because $\text{GraphSeg}^-$ tend to over-segment the scene as shown in \cref{fig:ablation}, which is supported by the corresponding lower $\text{IoU}_{sel}$ compared to GraphSeg with 3D graph edge addition and contraction. MaskClustering (MC) reaches good $d_{\text{chamfer}}$ but poor IoU, which is likely because it filters out many points, and it is challenging to find 3D relations in a sparse view setup. SAM3D provides reasonable performance in some scenes but not in all scenes. Compared to the two baseline methods, our method can robustly segment the scene and out-performs the very recent baselines significantly. Additionally, we also compare the precision between our method and the baseline methods. Methods such as MaskClustering aggressively filter out points which are uncertain, resulting in low recall, F1, and IoU measure, but do not adversely impact precision, lower Chamfer distance. As shown in \cref{tab:compare_precision}, MaskClustering produces reasonable results as measured by precision and also Chamfer distance. However, our method still outperforms the compared baselines, measured by all of the metrics.

\begin{figure}
\captionsetup[subfigure]{labelformat=empty}
\vspace{-1em}
    \centering
    \begin{subfigure}[t]{0.195\linewidth}
        \caption{Reconstruction}
        \centering
        \includegraphics[width=\linewidth]{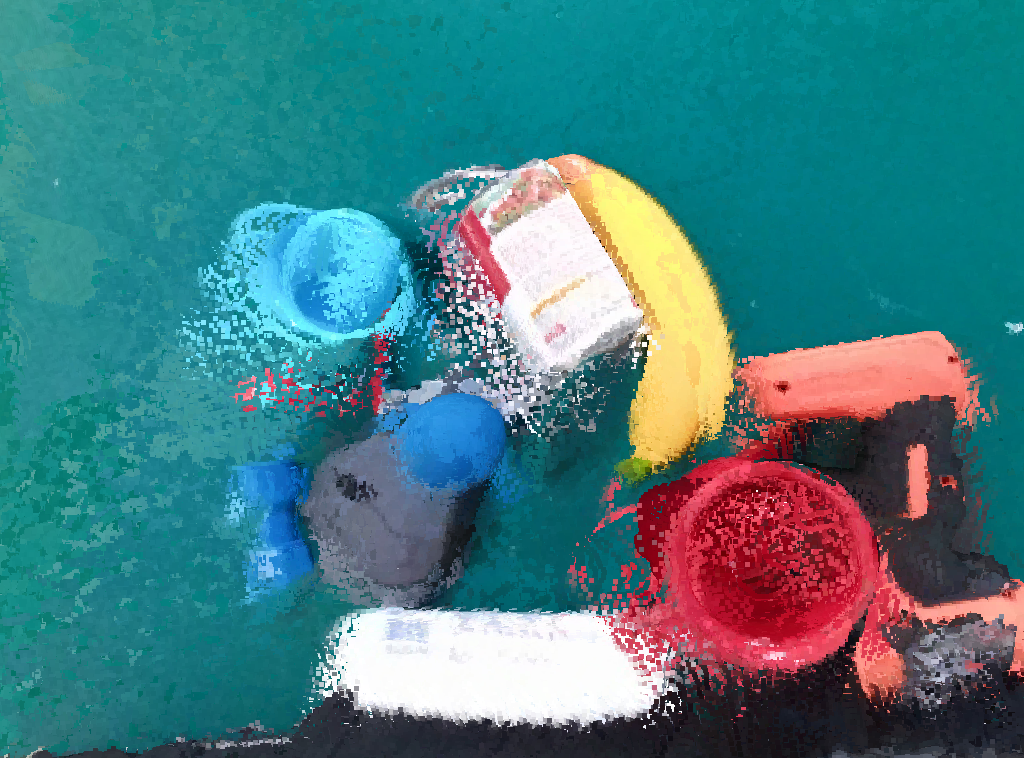}\\[0.5ex]
        \includegraphics[width=\linewidth]{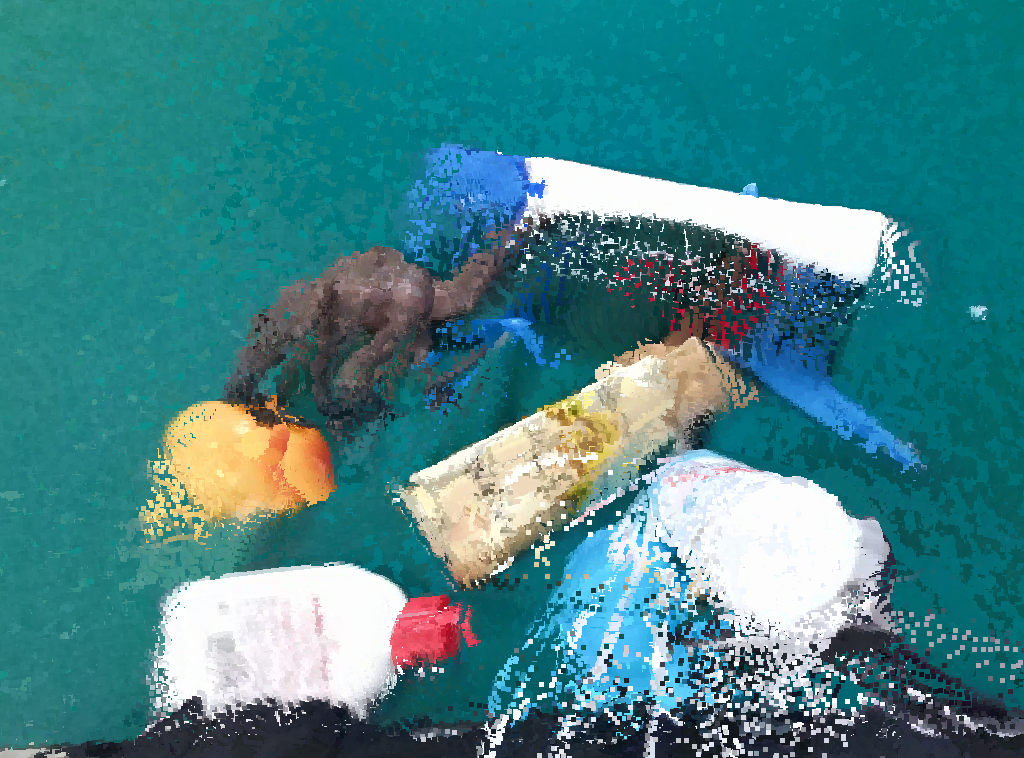}\\[0.5ex]
        \includegraphics[width=\linewidth]{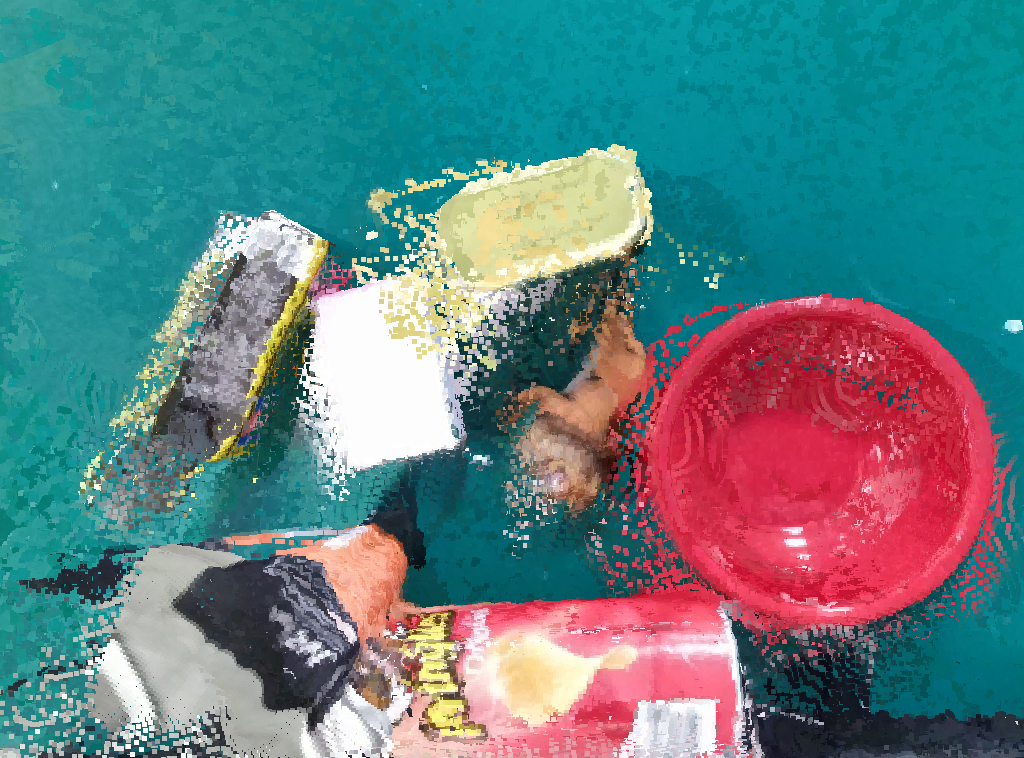}\\[0.5ex]
        \includegraphics[width=\linewidth]{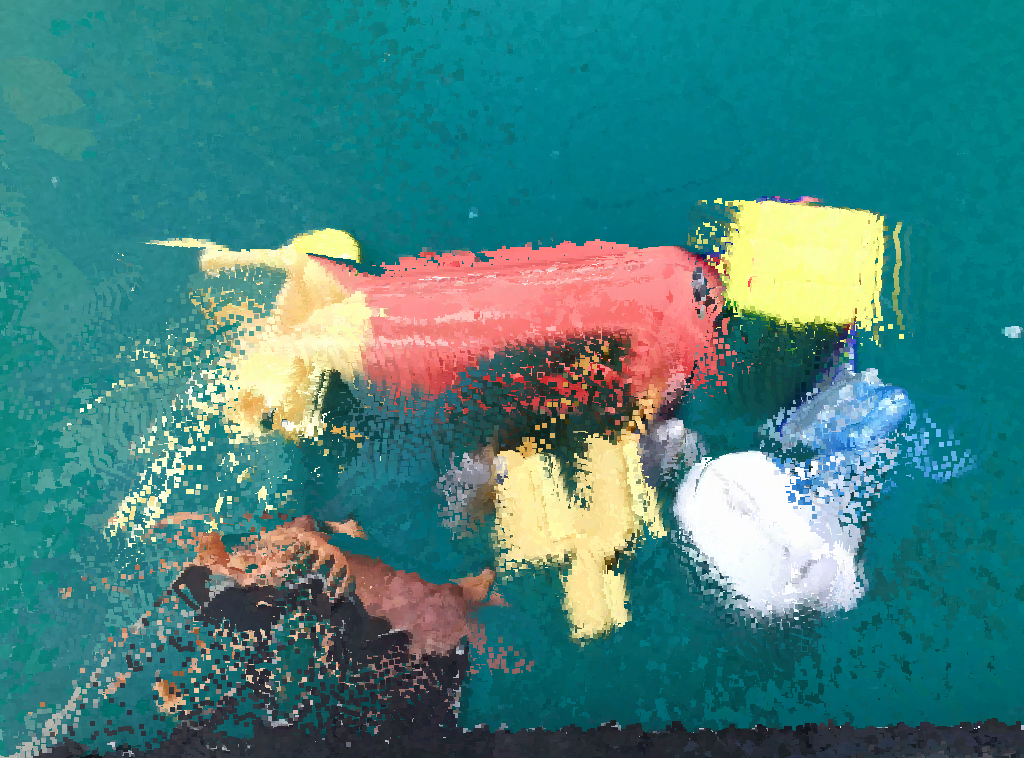}\\[0.5ex]
        \includegraphics[width=\linewidth]{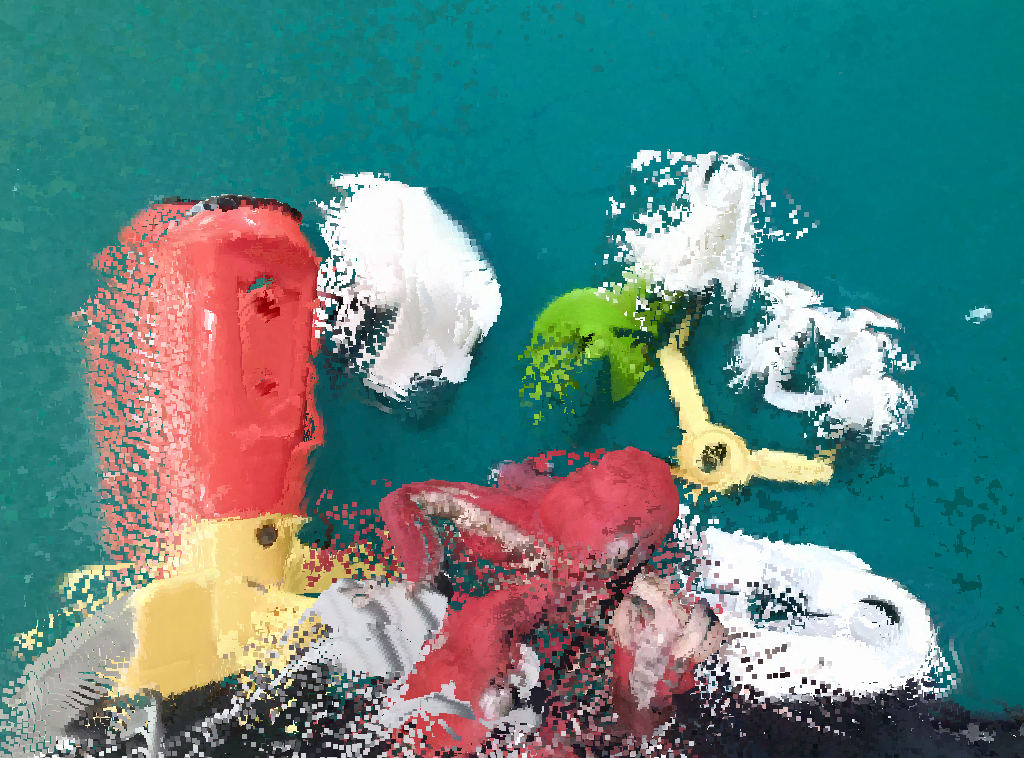}
    \end{subfigure}\hfill
    \begin{subfigure}[t]{0.195\linewidth}
        \caption{SAM3D}
        \centering
        \includegraphics[width=\linewidth]{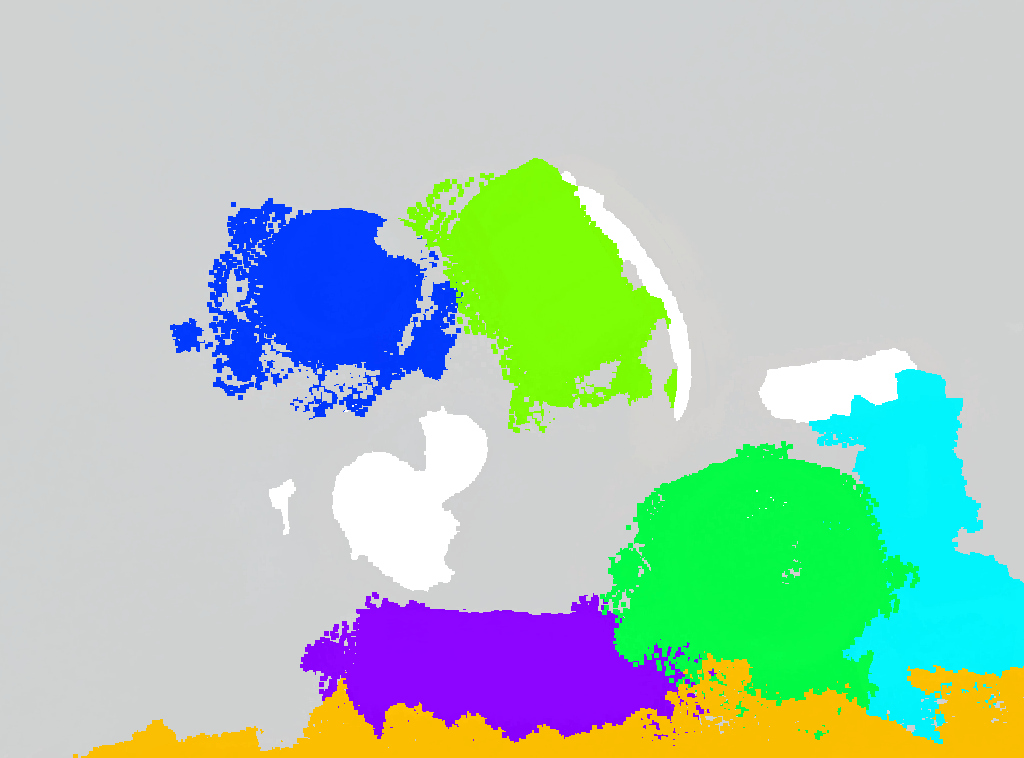}\\[0.5ex]
        \includegraphics[width=\linewidth]{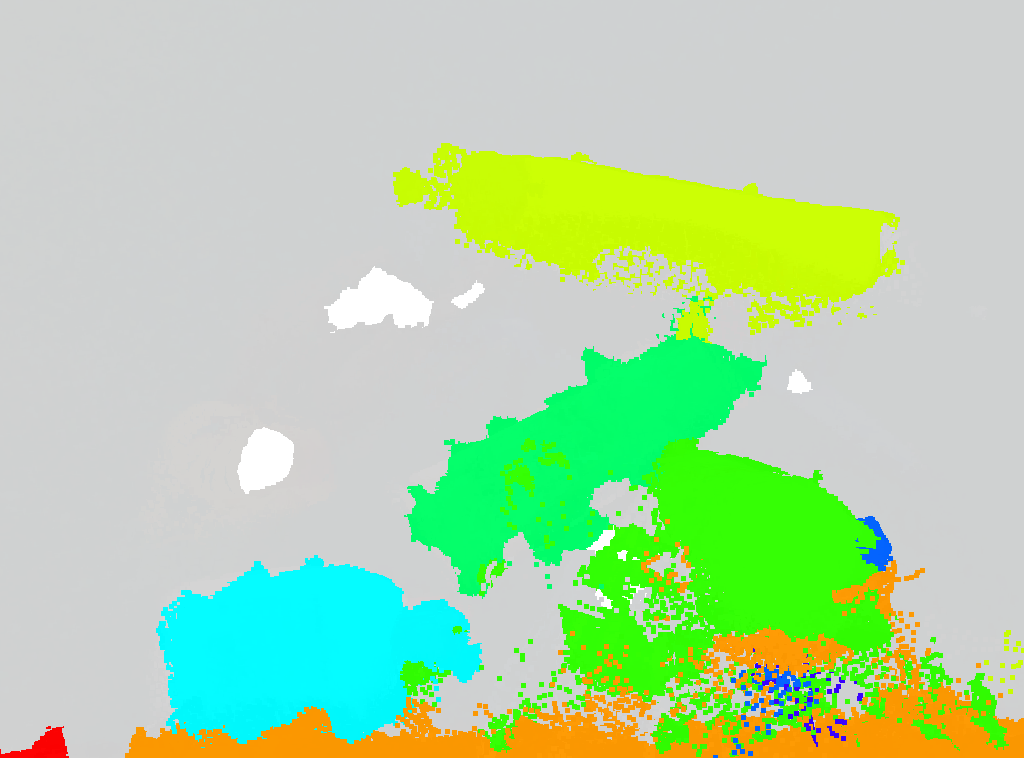}\\[0.5ex]
        \includegraphics[width=\linewidth]{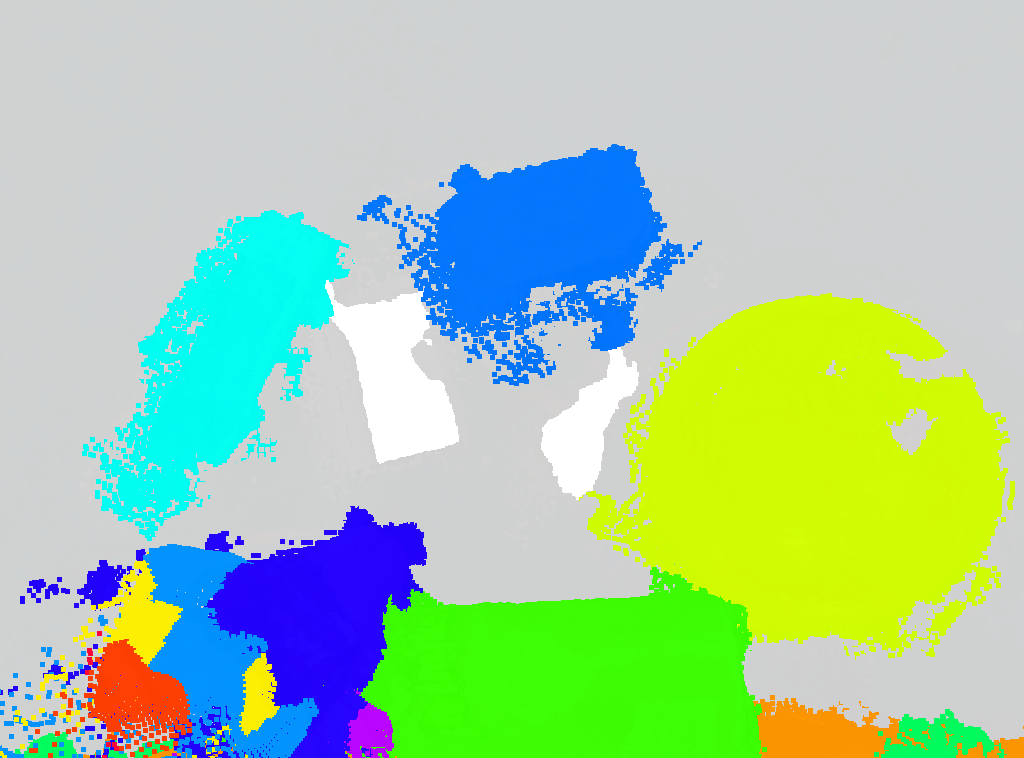}\\[0.5ex]
        \includegraphics[width=\linewidth]{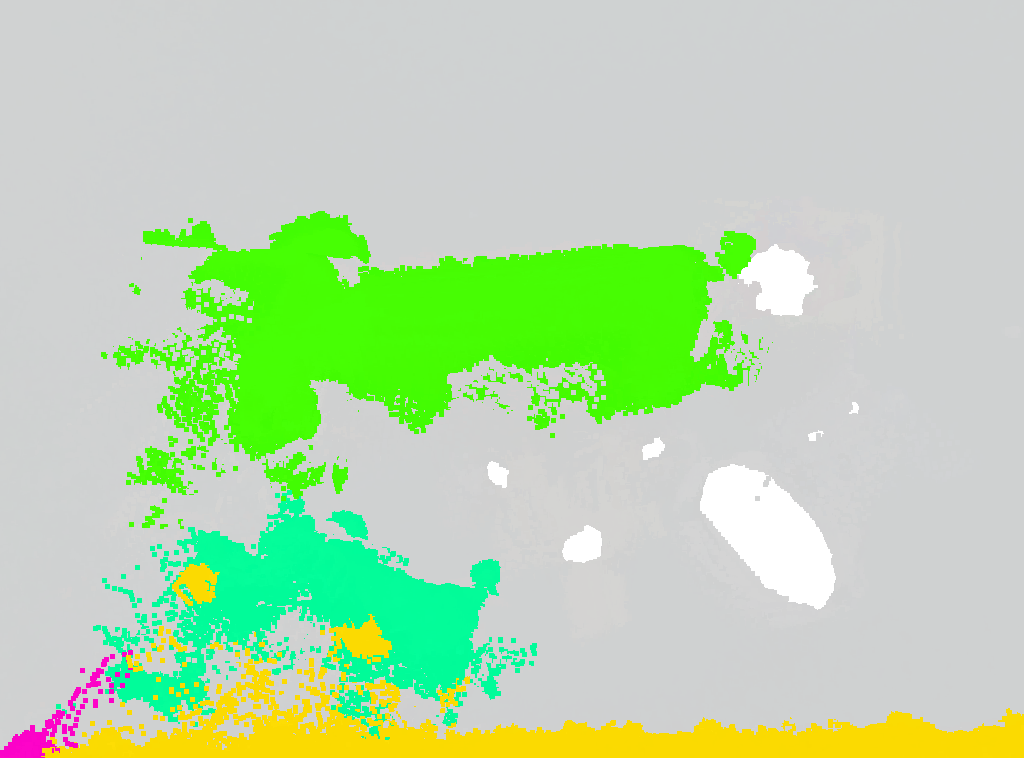}\\[0.5ex]
        \includegraphics[width=\linewidth]{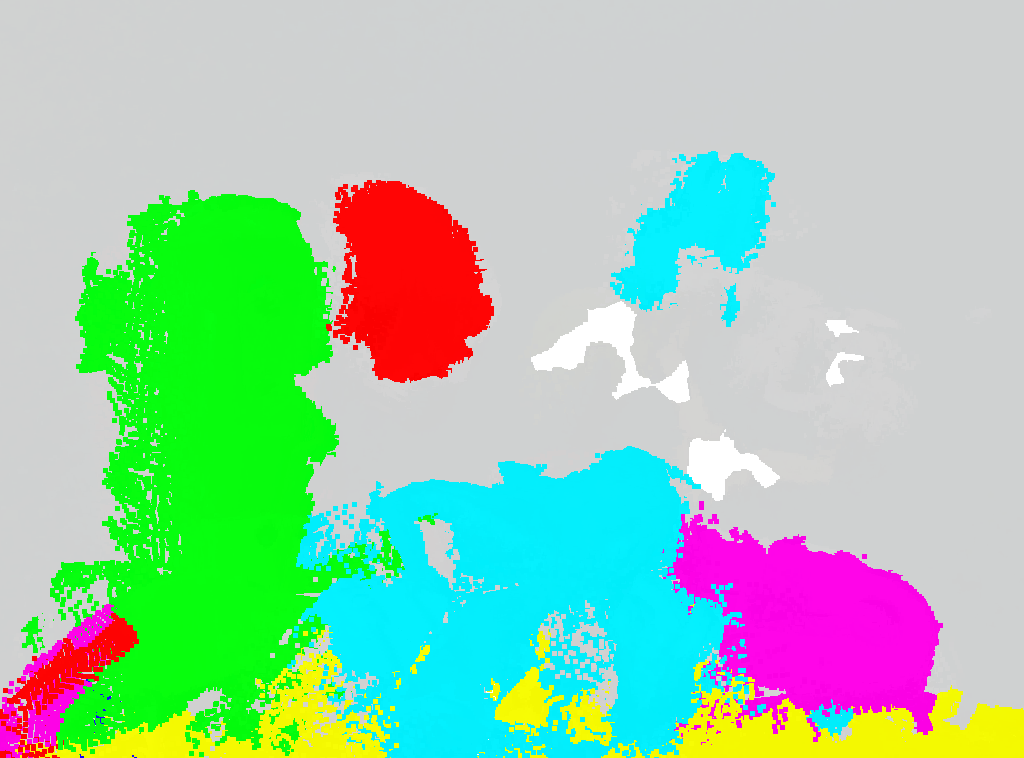}
    \end{subfigure}\hfill
    \begin{subfigure}[t]{0.195\linewidth}
        \caption{MC}
        \centering
        \includegraphics[width=\linewidth]{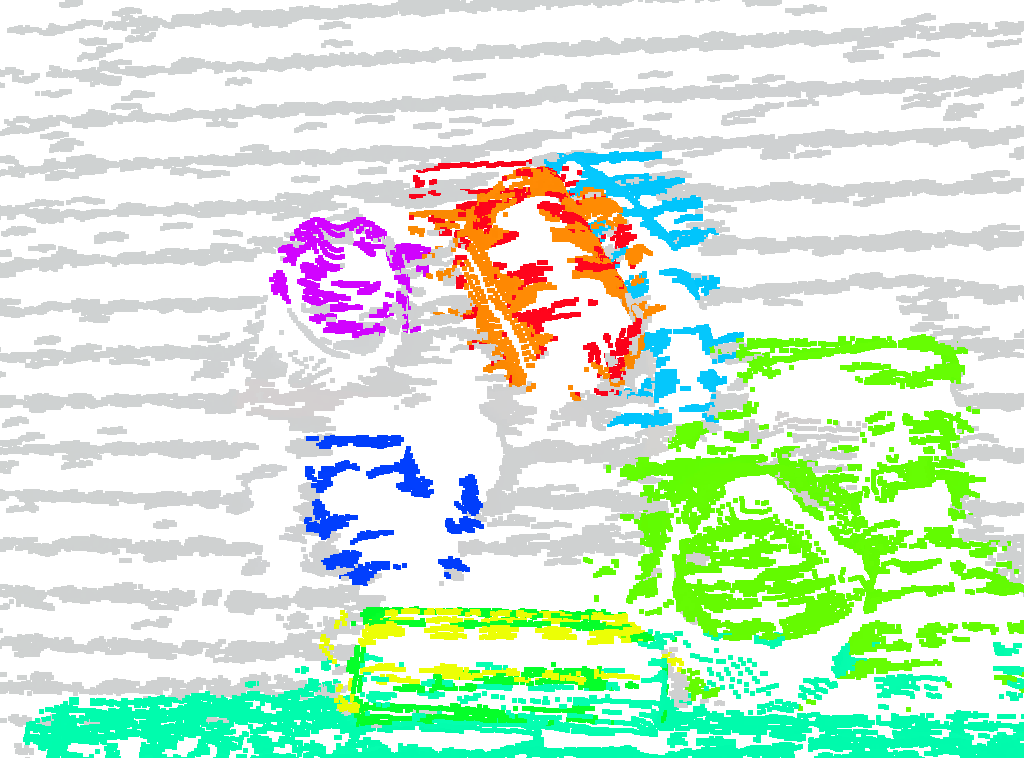}\\[0.5ex]
        \includegraphics[width=\linewidth]{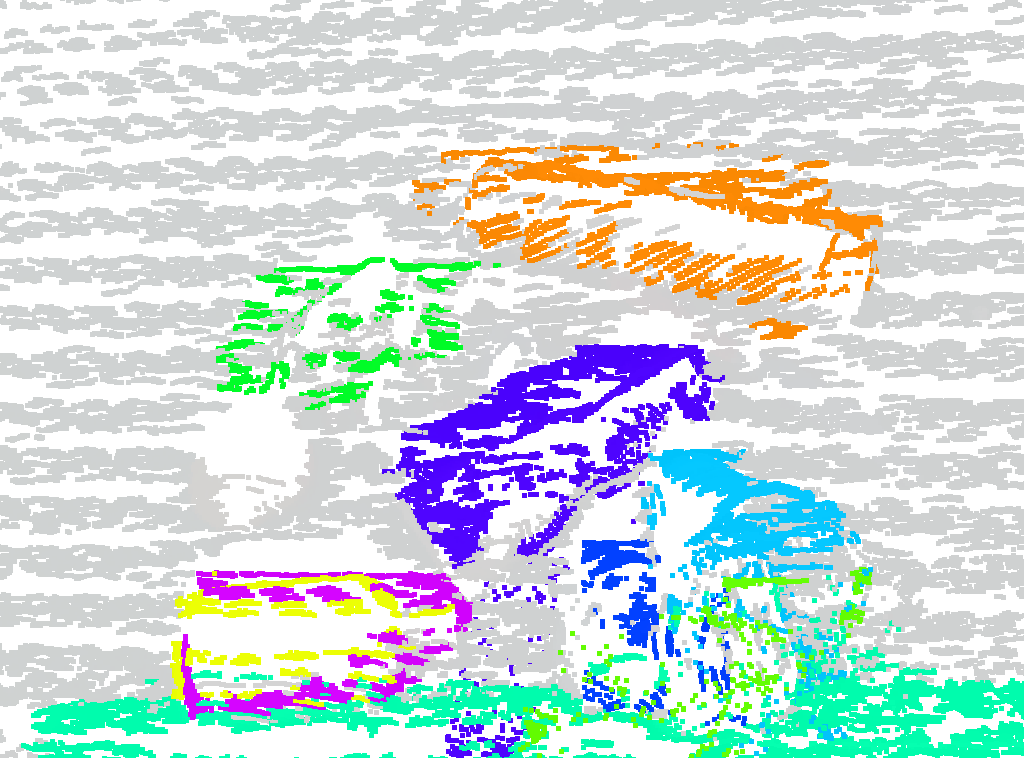}\\[0.5ex]
        \includegraphics[width=\linewidth]{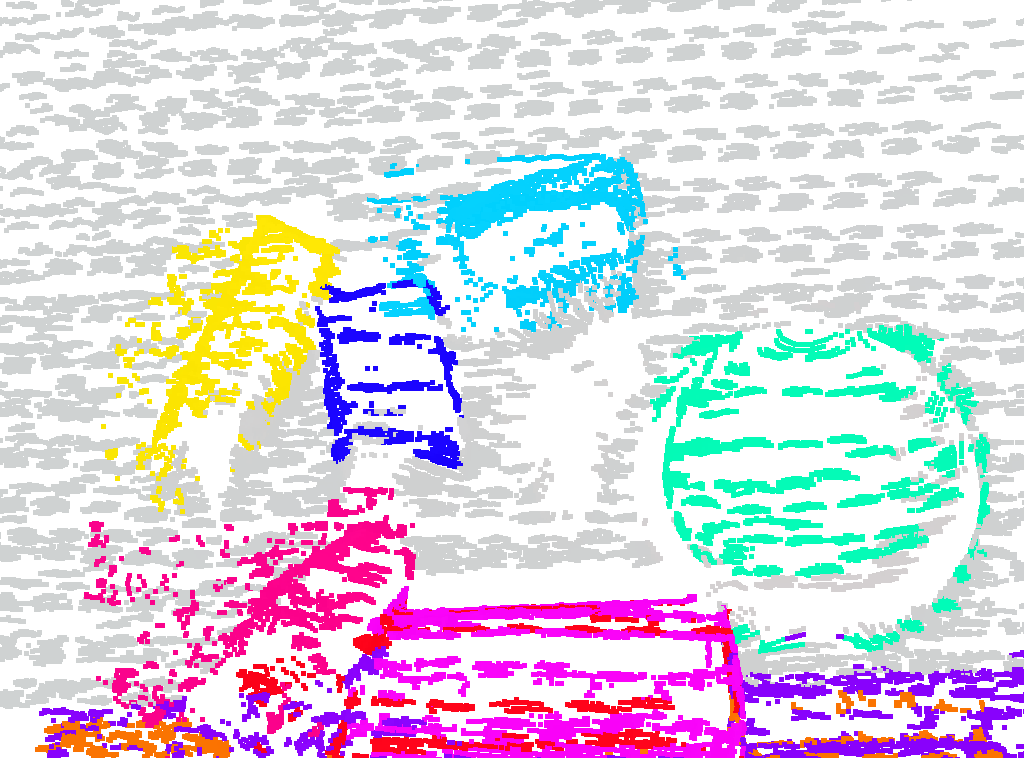}\\[0.5ex]
        \includegraphics[width=\linewidth]{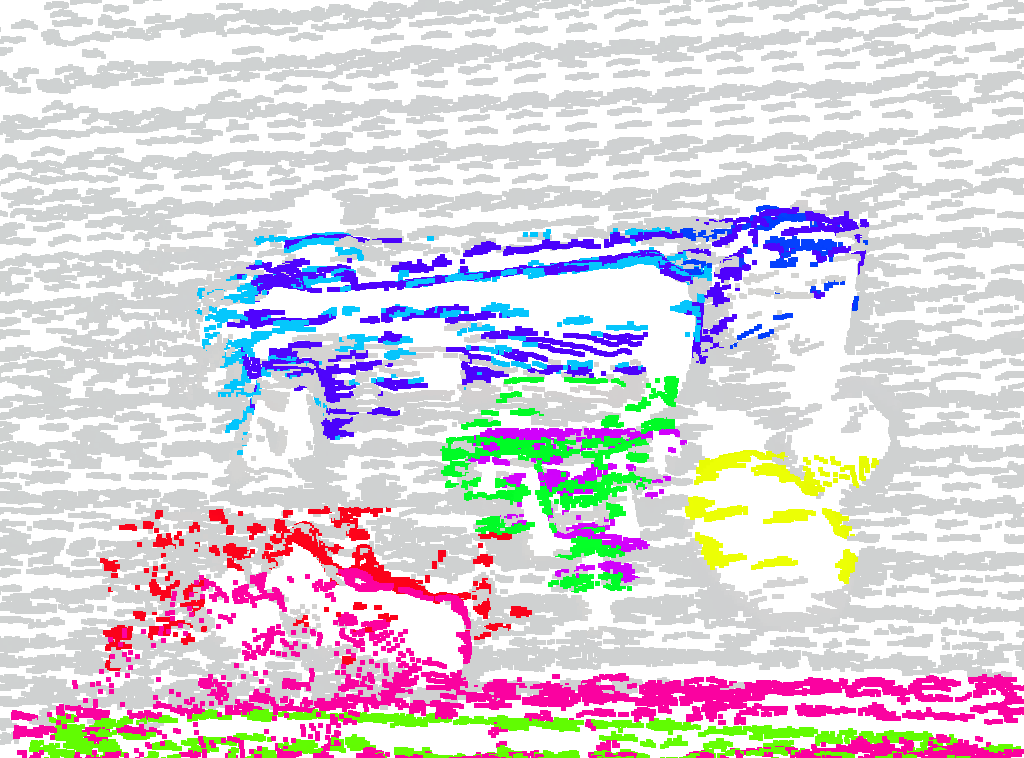}\\[0.5ex]
        \includegraphics[width=\linewidth]{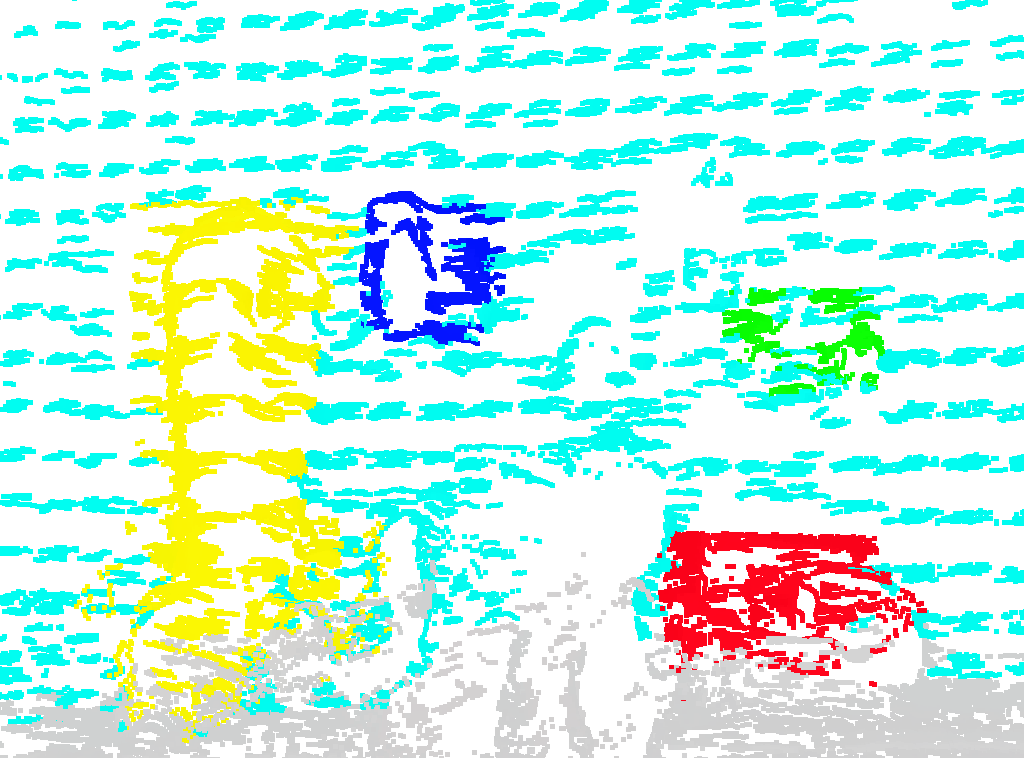}
    \end{subfigure}\hfill
    \begin{subfigure}[t]{0.195\linewidth}
        \caption{\textbf{Ours}}
        \centering
        \includegraphics[width=\linewidth]{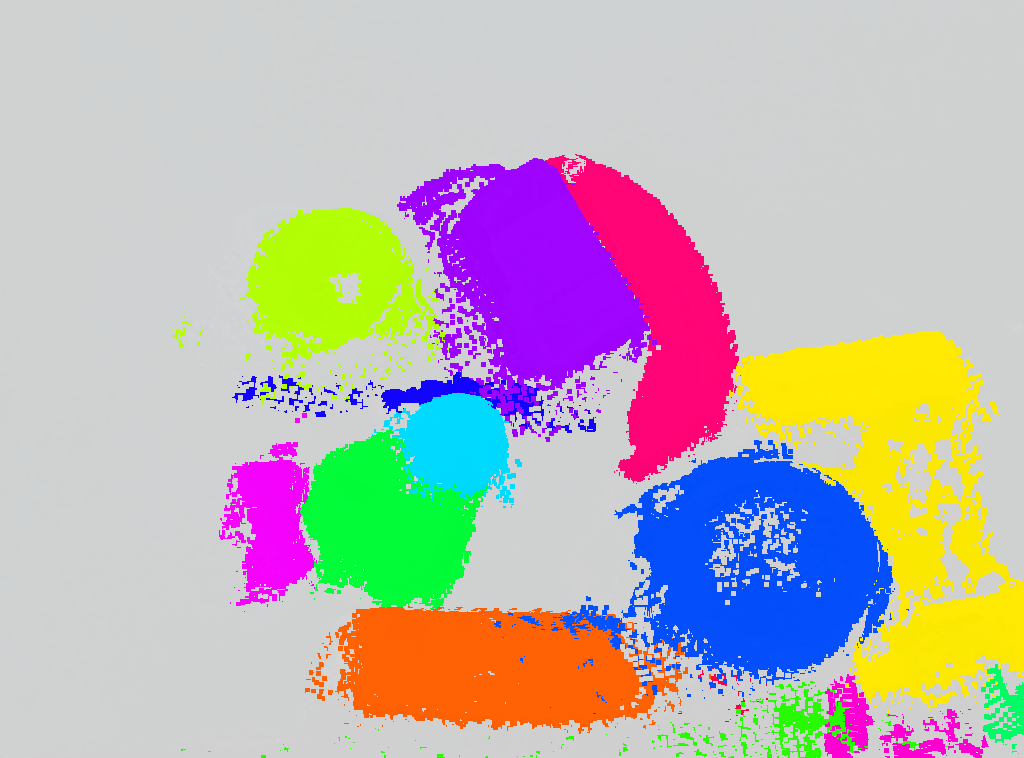}\\[0.5ex]
        \includegraphics[width=\linewidth]{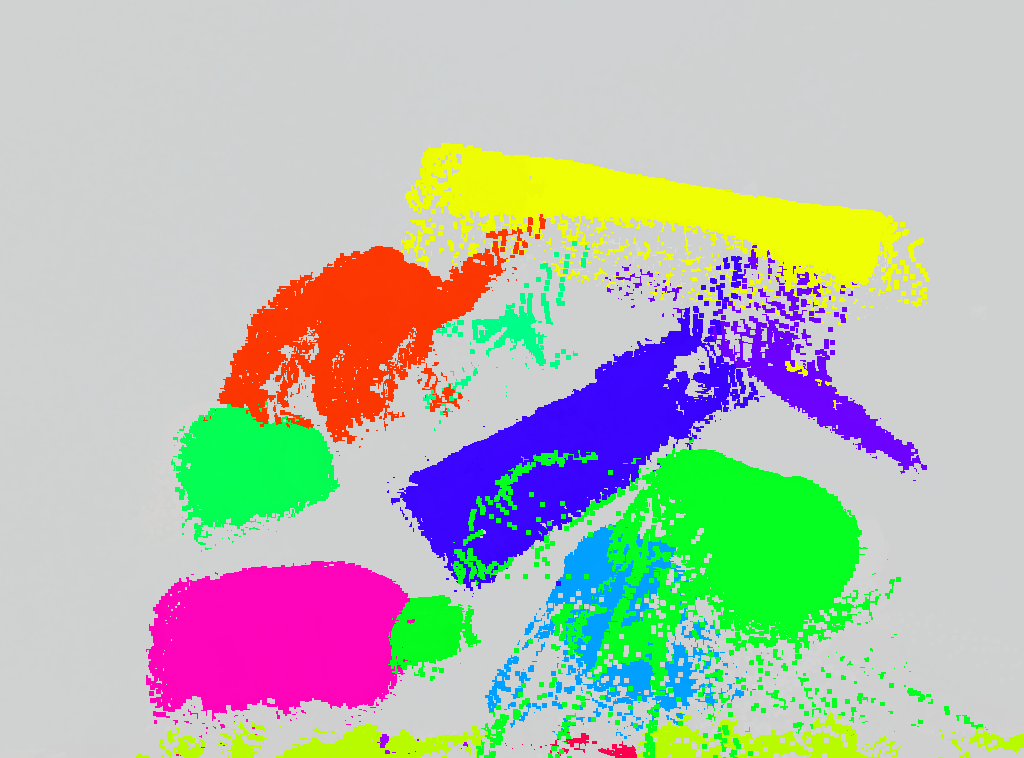}\\[0.5ex]
        \includegraphics[width=\linewidth]{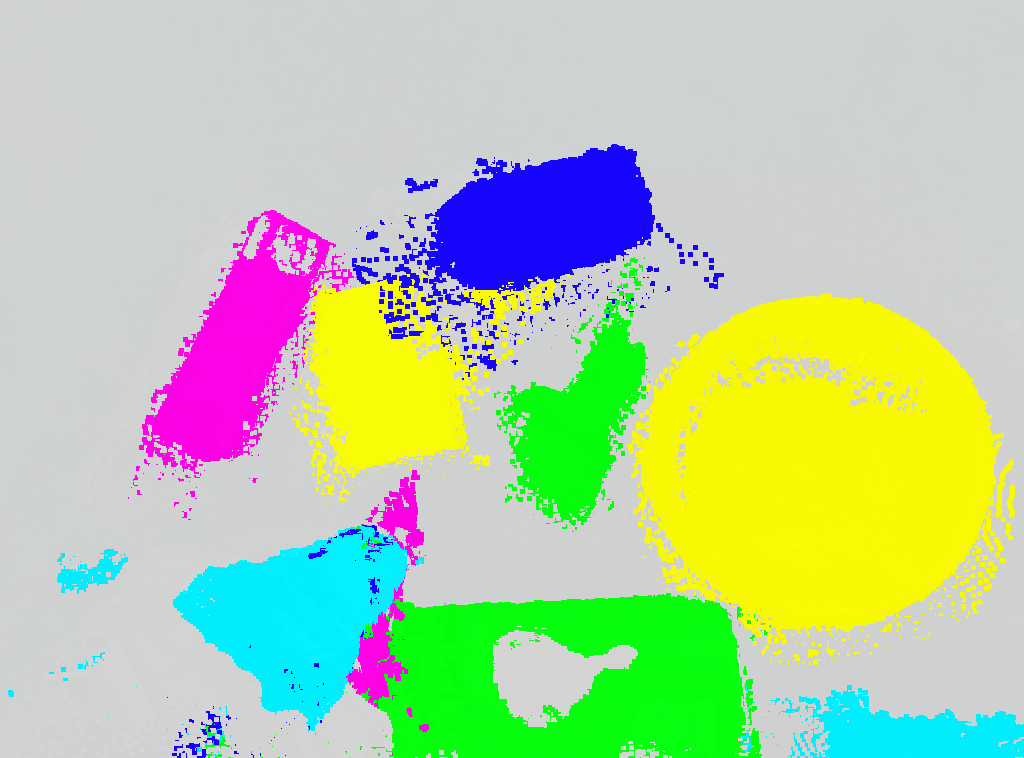}\\[0.5ex]
        \includegraphics[width=\linewidth]{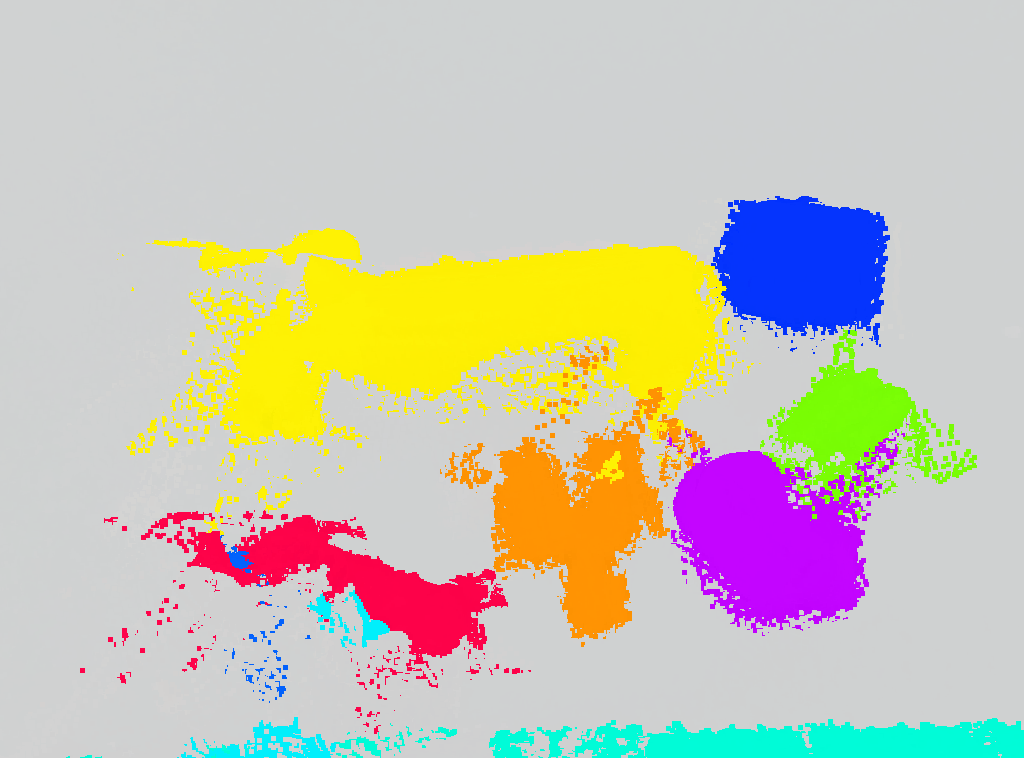}\\[0.5ex]
        \includegraphics[width=\linewidth]{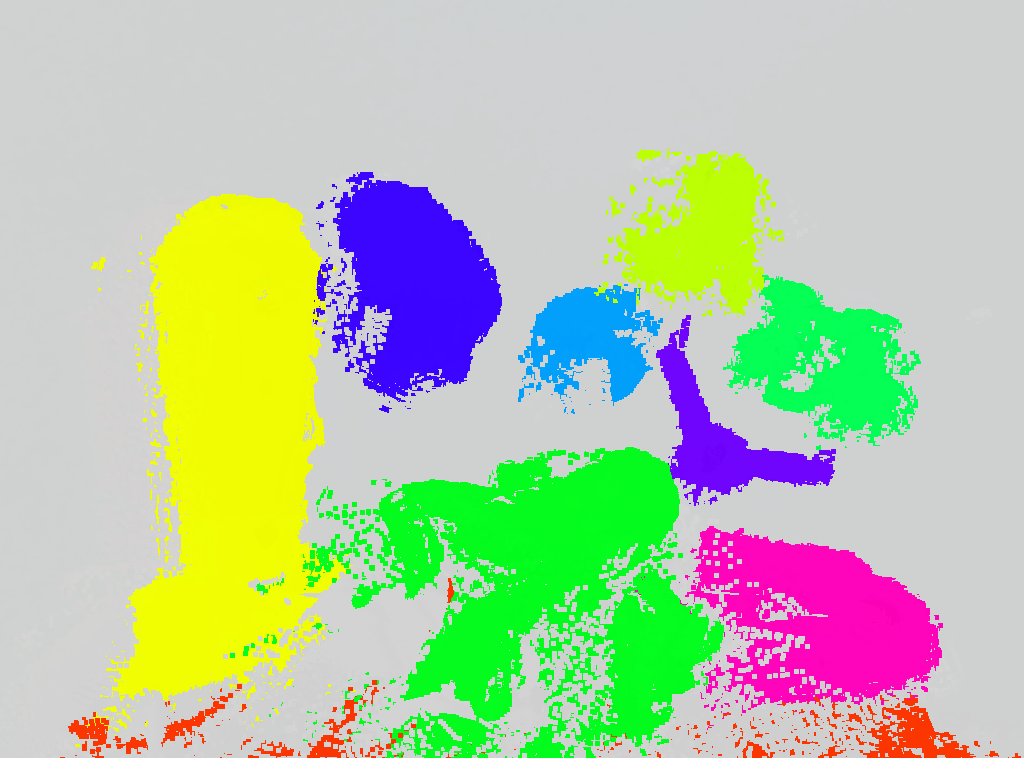}
    \end{subfigure}\hfill
    \begin{subfigure}[t]{0.195\linewidth}
        \caption{GT}
        \centering
        \includegraphics[width=\linewidth]{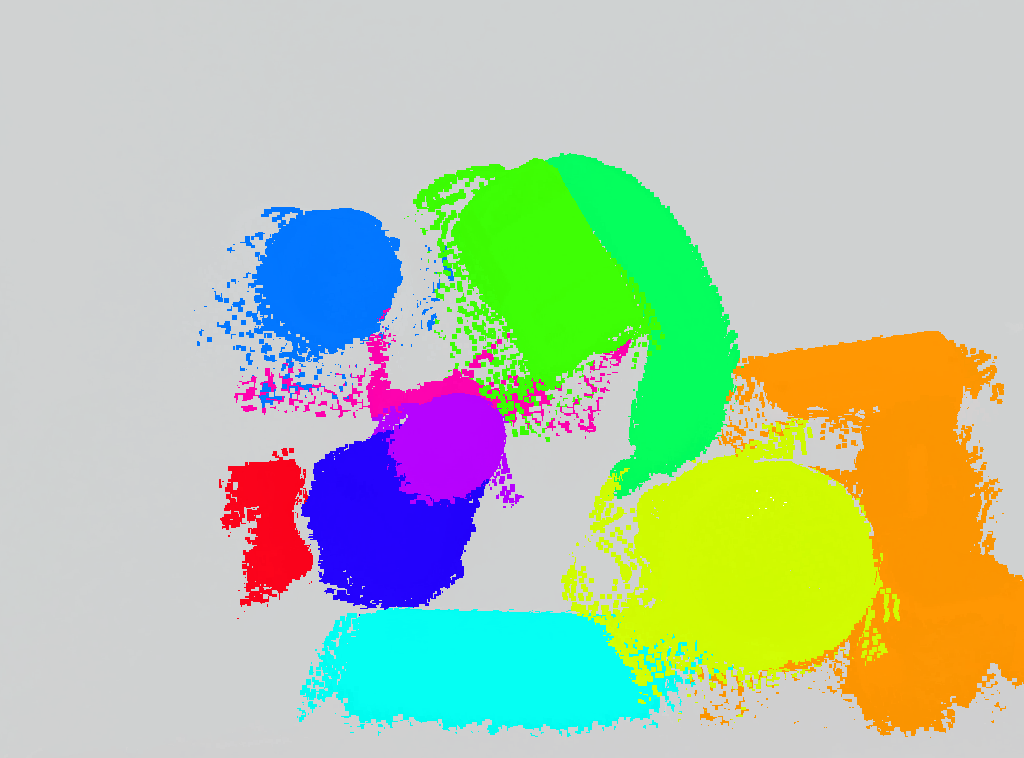}\\[0.5ex]
        \includegraphics[width=\linewidth]{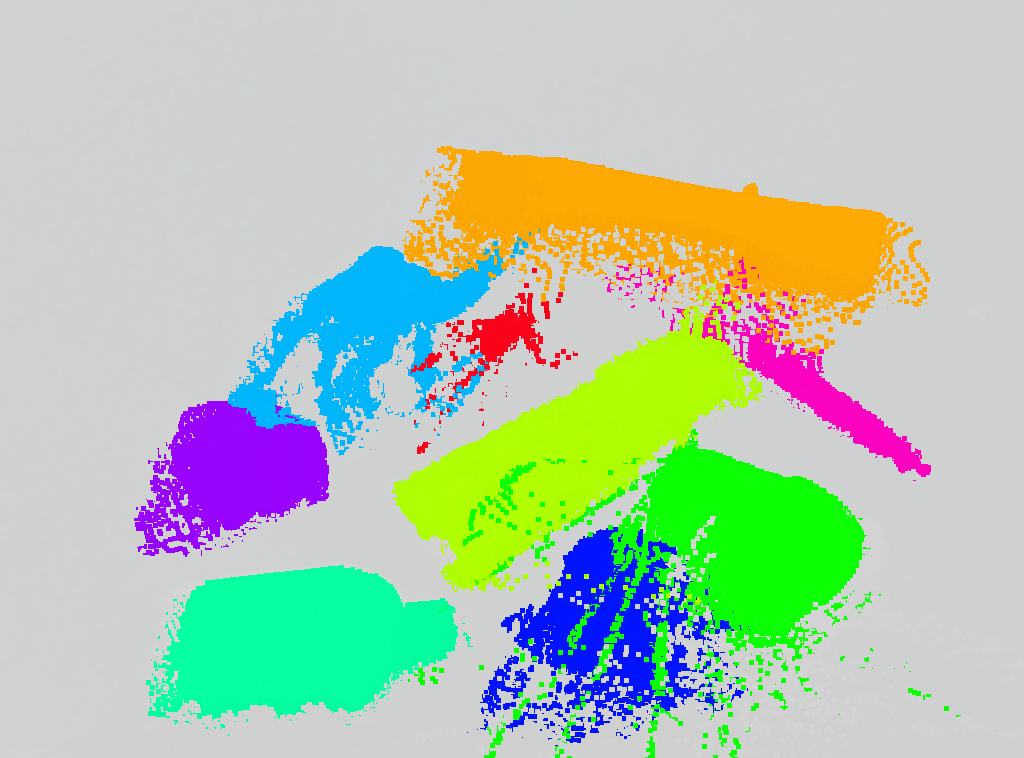}\\[0.5ex]
        \includegraphics[width=\linewidth]{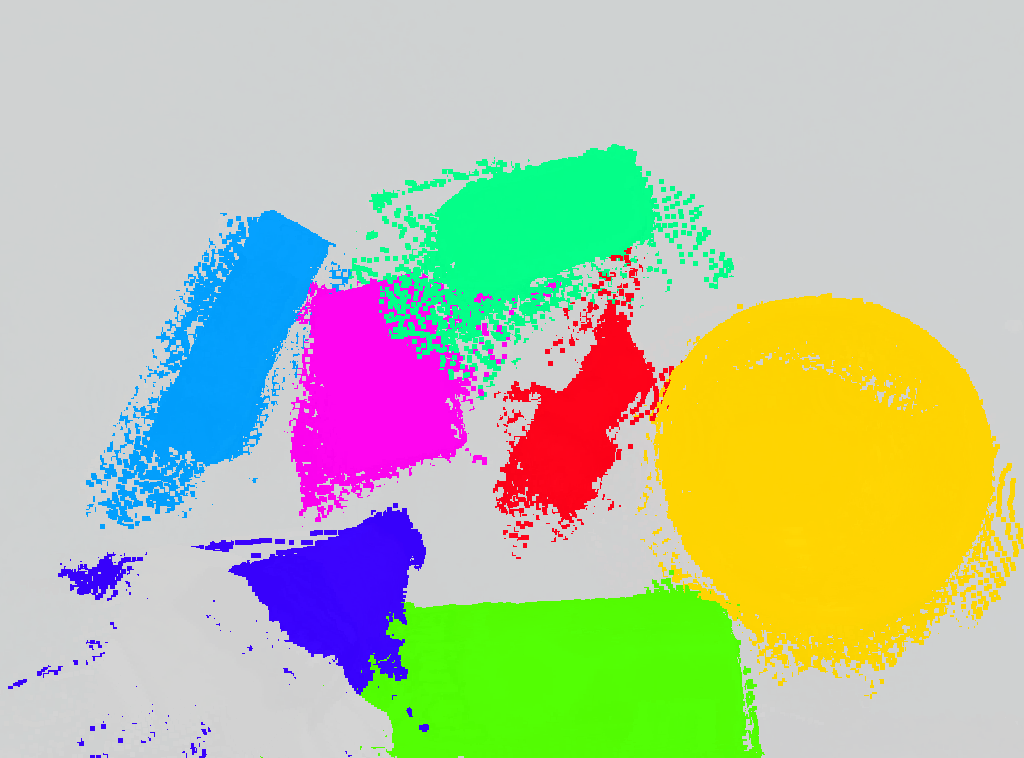}\\[0.5ex]
        \includegraphics[width=\linewidth]{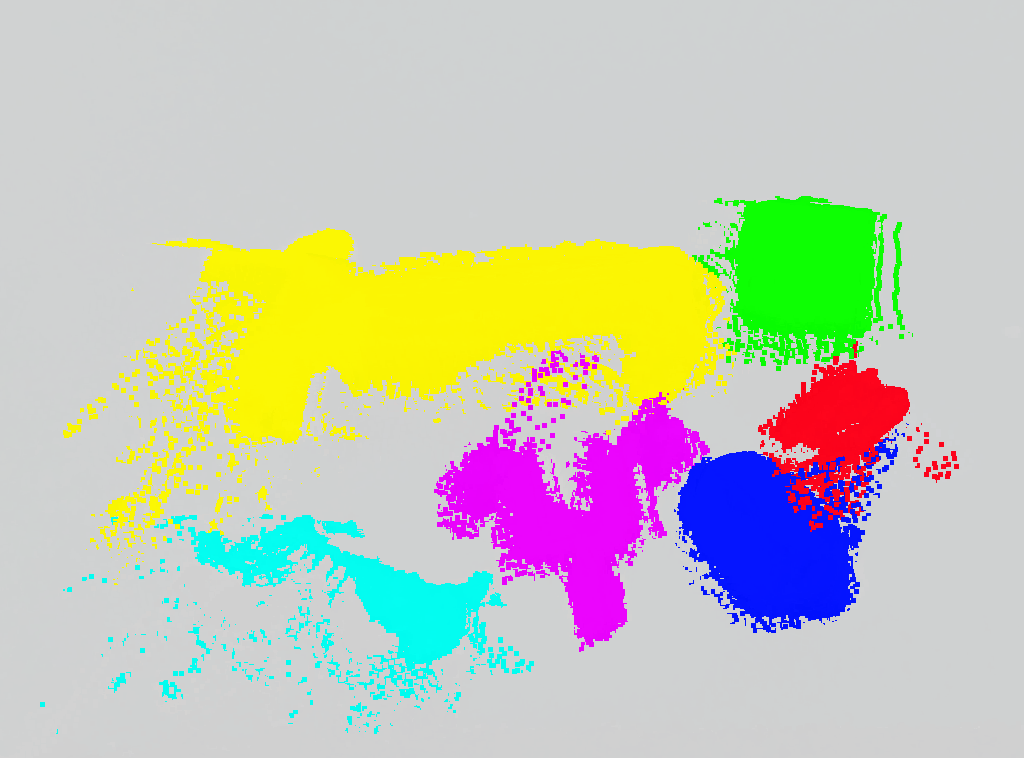}\\[0.5ex]
        \includegraphics[width=\linewidth]{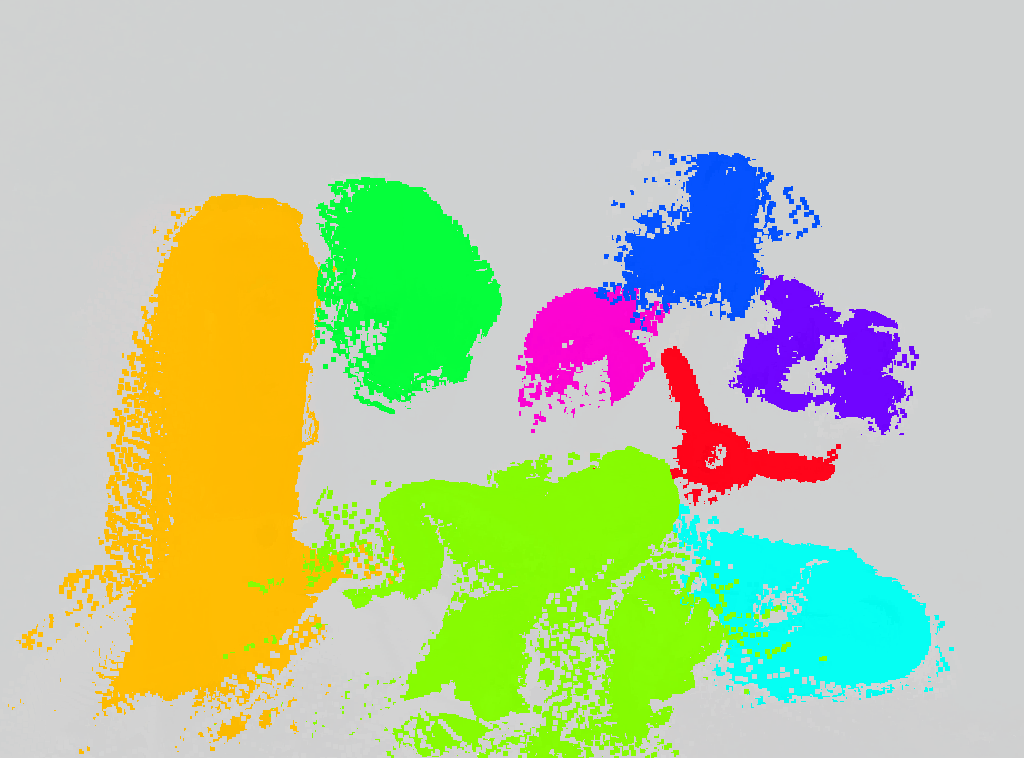}
    \end{subfigure}
    
    \caption{Qualitative comparison of our method against the two baseline methods \cite{yang2023sam3d, yan2024maskclustering} on GraspNet-1B \cite{fang2020graspnet}. MaskClustering aggressively filters out many points as a part of its filtering and point cloud refinement, producing significantly sparser 3D segmented point clouds.}
    \label{fig:qual_compare}
\end{figure}

\begin{figure}[t]
  \centering
  \begin{subfigure}[t]{0.24\textwidth}
    \centering
    \includegraphics[width=0.42\textwidth]{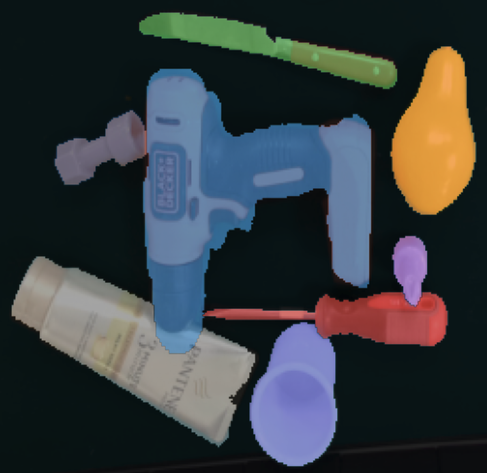}%
    \includegraphics[width=0.56\textwidth]{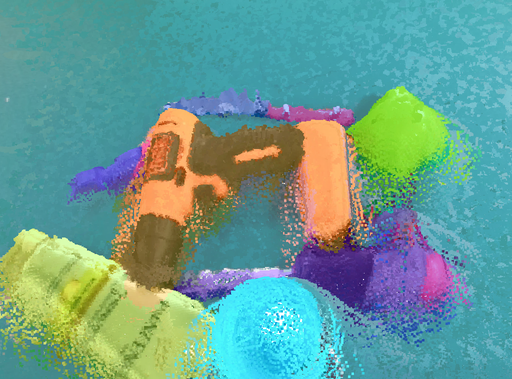}
    \caption{Ground Truth}
  \end{subfigure}%
  \begin{subfigure}[t]{0.24\textwidth}
    \centering
    \includegraphics[width=0.42\textwidth]{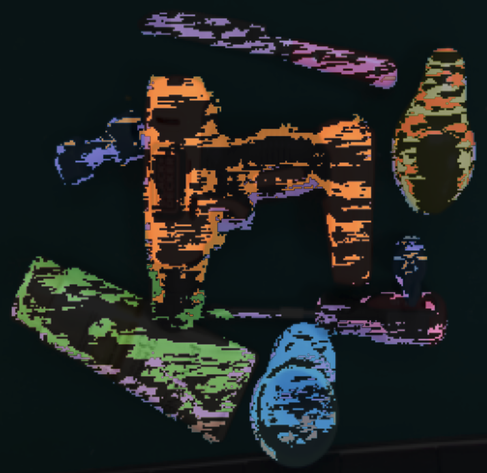}%
    \includegraphics[width=0.56\textwidth]{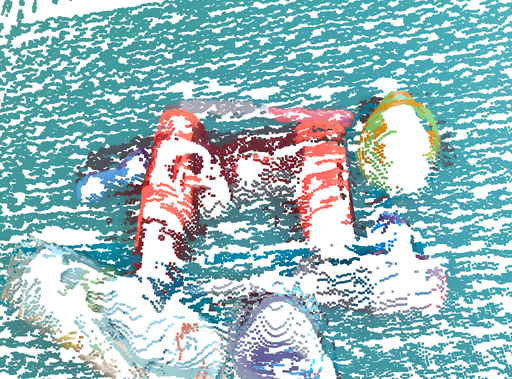}
    \caption{MaskClustering}
  \end{subfigure}%
  
  \begin{subfigure}[t]{0.24\textwidth}
    \centering
    \includegraphics[width=0.42\textwidth]{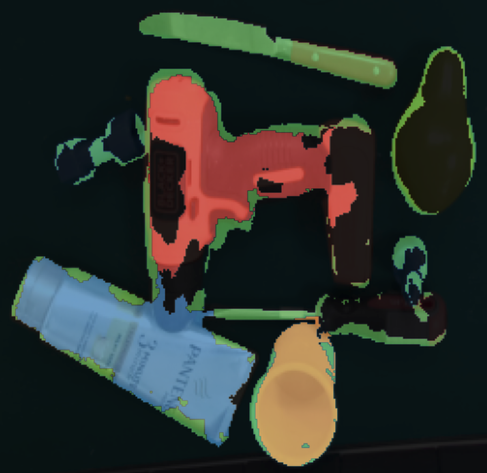}%
    \includegraphics[width=0.56\textwidth]{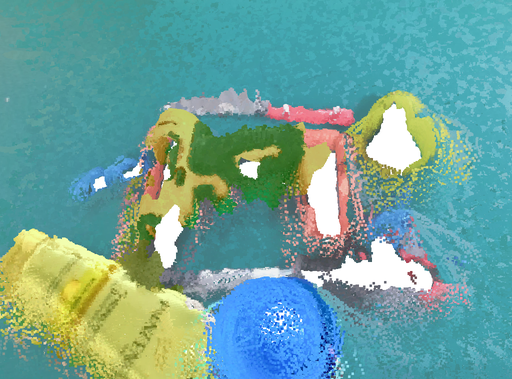}
    \caption{SAM3D}
  \end{subfigure}%
  \begin{subfigure}[t]{0.24\textwidth}
    \centering
    \includegraphics[width=0.42\textwidth]{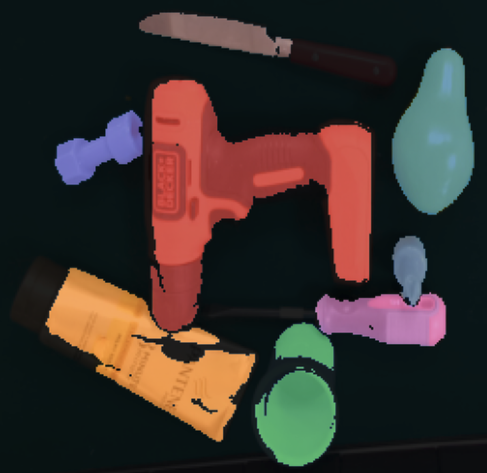}%
    \includegraphics[width=0.56\textwidth]{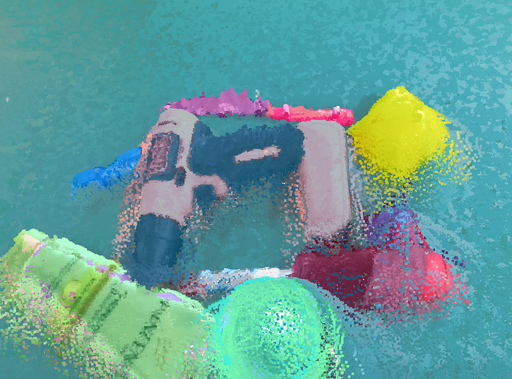}
    \caption{GraphSeg}
  \end{subfigure}
  \caption{For each method, its segmentation output is visualized in 2D (on the left) and the corresponding segmented point cloud on the right. We can see that both SAM3D and MaskClustering suffer from over-segmentation, which caused objects in the segmented 3D points to be noisy, missing, or incomplete, while GraphSeg shows its robustness against over-segmentation to ensure objects are intact.}\label{fig:robustness}
\end{figure}

To qualitatively assess our method, we visualize the result of our method against MaskClustering \cite{yan2024maskclustering} and SAM3D \cite{yang2023sam3d} in GraspNet-1B \cite{fang2020graspnet} in \cref{fig:qual_compare}. As shown, our method avoids the under-segmentation, displayed by SAM3D, by using 2D correspondence to first get an initial fine-grained segmentation. This reduces aggressive merging, when relying heavily 3D structure as is done by SAM3D. MaskClustering has an over-segmentation issue in the experiment, which is likely caused by the challenging sparse multi-view setup where the spatial relationship is harder to derive. Our method solves this by fusing 2D correspondence and 3D correspondence information to build additional edges and contracting the segmentation graph more robustly. An illustration of GraphSeg relative to comparison methods is also shown in \cref{fig:robustness}.

\begin{figure}[t]
  \centering
  \begin{subfigure}[b]{0.24\linewidth}
    \centering
    \includegraphics[width=\linewidth]{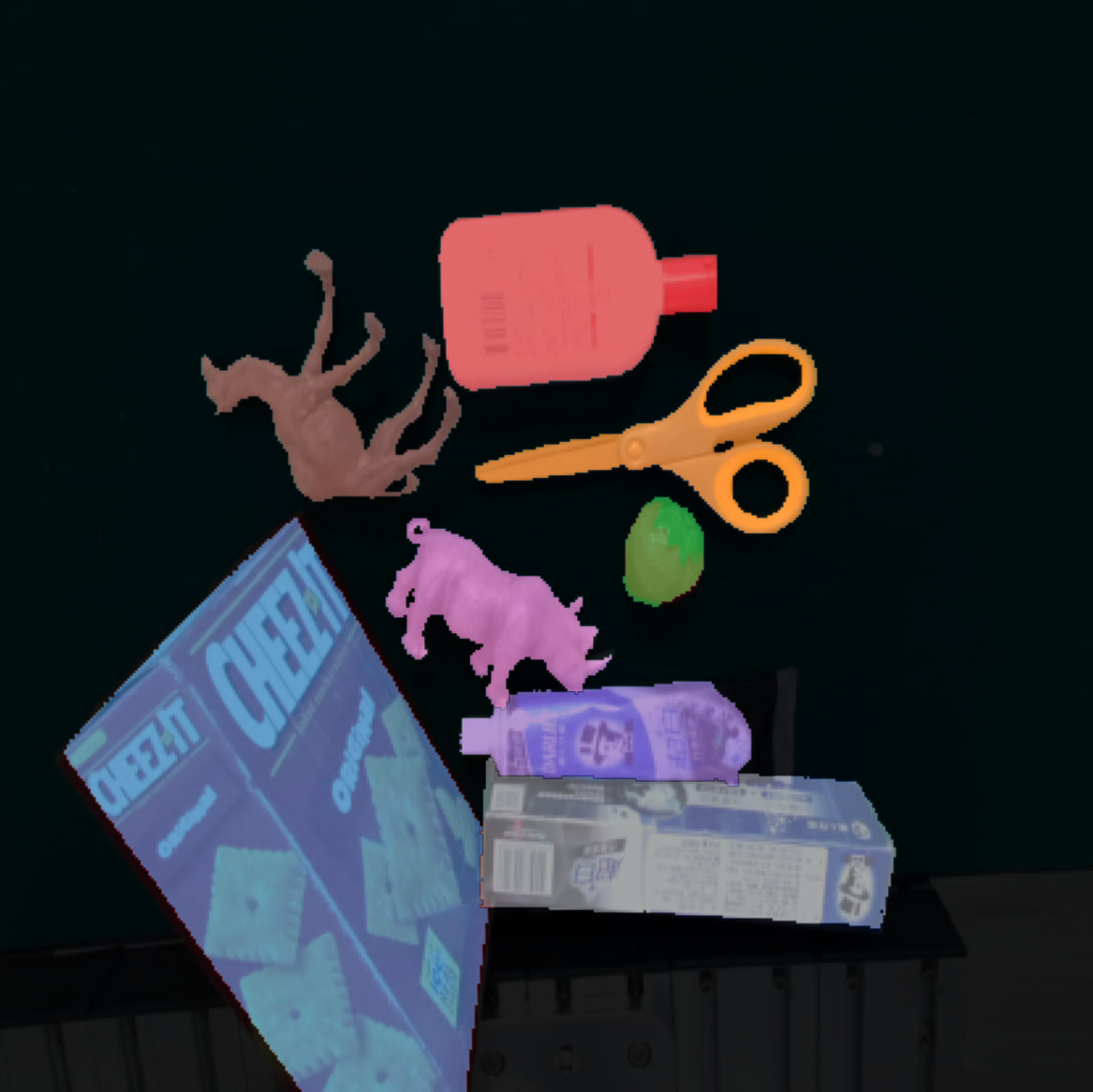}
    \caption{Ground Truth}
  \end{subfigure}%
  \begin{subfigure}[b]{0.24\linewidth}
    \centering
    \includegraphics[width=\linewidth]{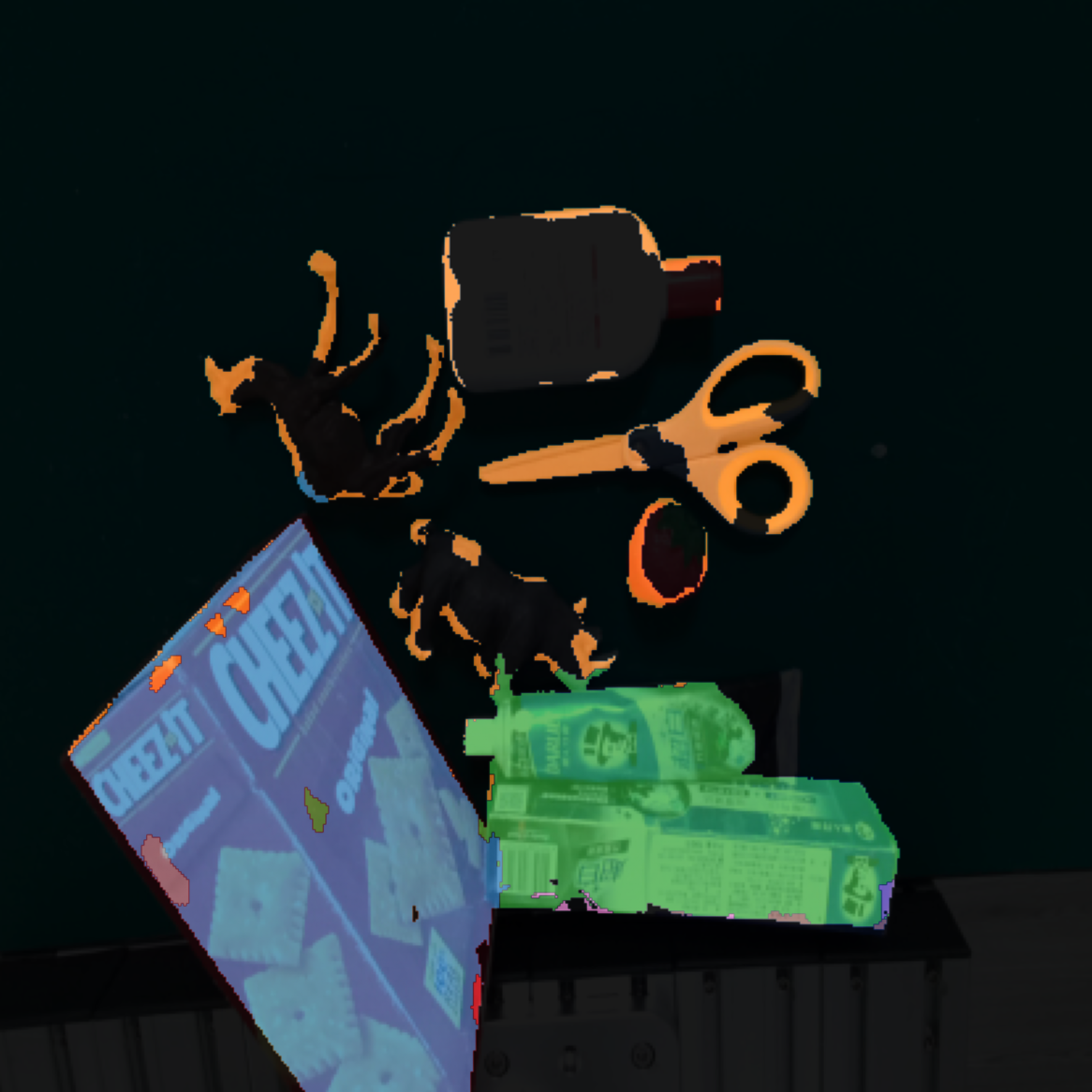}
    \caption{SAM3D}
  \end{subfigure}%
  \begin{subfigure}[b]{0.24\linewidth}
    \centering
    \includegraphics[width=\linewidth]{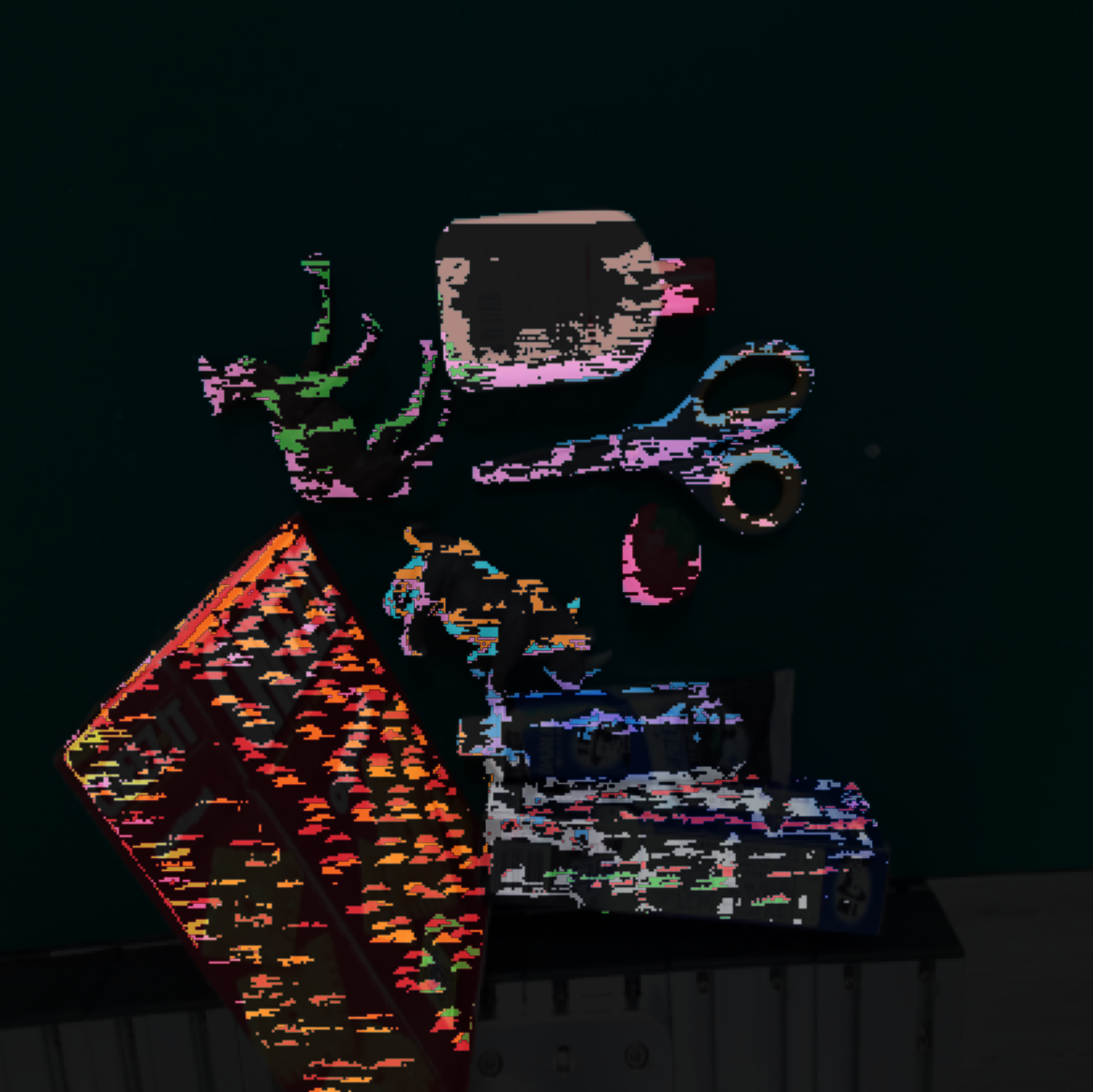}
    \caption{MC}
  \end{subfigure}%
  \begin{subfigure}[b]{0.24\linewidth}
    \centering
    \includegraphics[width=\linewidth]{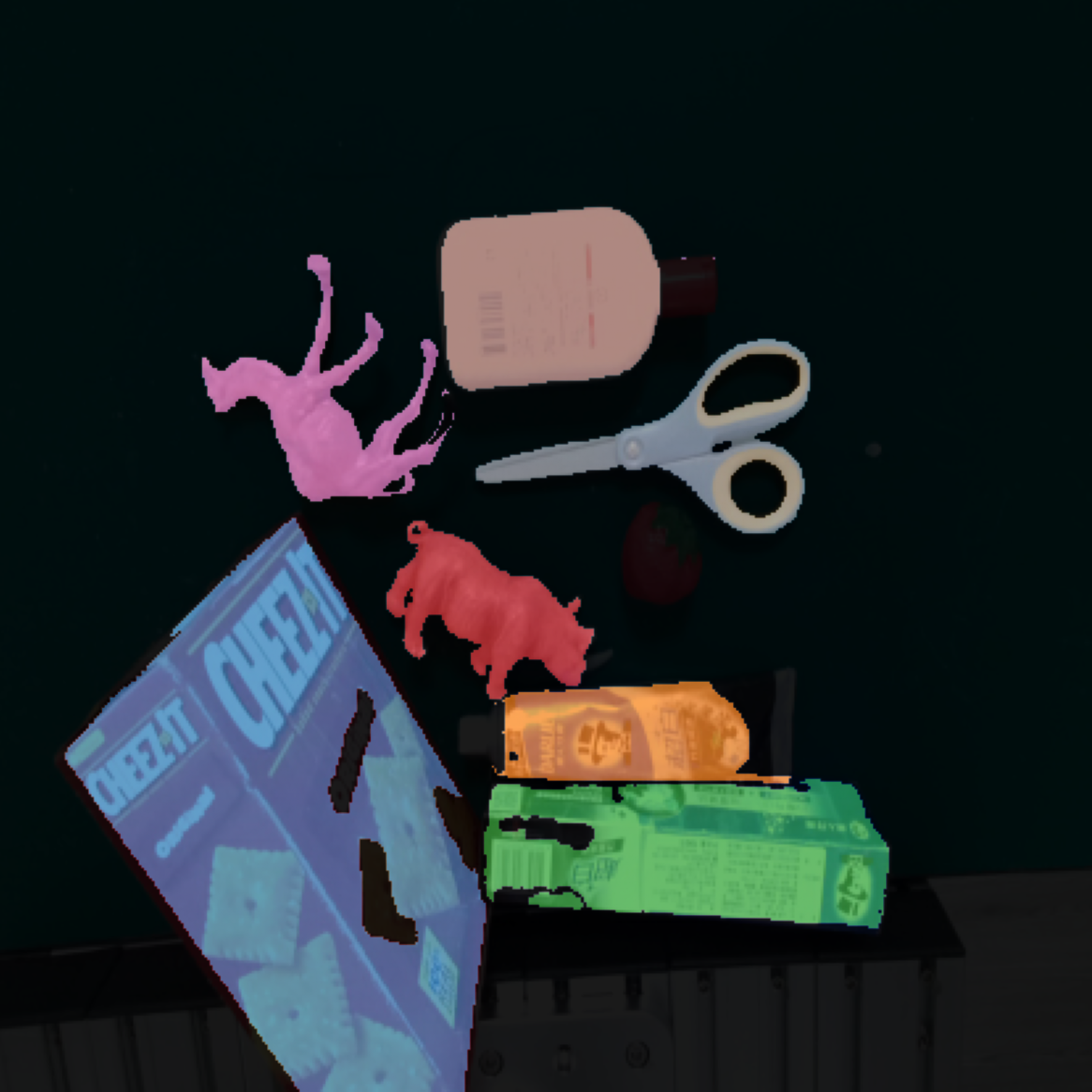}
    \caption{GraphSeg}
  \end{subfigure}
  \caption{We observe that our GraphSeg accounts for nearly all of the available pixels for the non-background object pixels considered, whereas SAM3D and MaskClustering alternatives filter away many pixels, leading to lower pixel utility.}
  \label{fig:pix_util}
\end{figure}

\begin{table}[t]
  \centering
  \begin{tabular}{lccc}
    \toprule
    Pixel Utility & MaskClustering & SAM3D & GraphSeg (ours) \\ \midrule
    Median & 0.0268 & 0.8648 & \textbf{0.9305} \\ 
    Mean   & 0.0267 & 0.8506 & \textbf{0.9263} \\ 
    \bottomrule
  \end{tabular}
  \caption{Here, we compare the mean and median of our method's pixel utility against baseline methods, over all our available datasets.}\label{tab:pixel_utility}
\end{table}

\subsection{GraphSeg Produces Solutions with High Pixel Utility}
Downstream robotics tasks generally benefit from denser and more complete point clouds. The process of constructing 3D representation, including the 3D foundation model, along with previous methods to construct consistent segmentation masks, filters out pixels for which the models are not confident. This is particularly true for MaskClustering, which relies on aggressively filtering out uncertain points. More 2D pixels being utilized to project into 3D space provides us with a denser point cloud. Here, we seek to measure, for non-background pixels across images, how many are utilized by the resulting masks. The pixel utility is therefore computed by taking the number of non-background pixels identified by each method over the actual number of non-background pixels in ground truth across images, shown in \cref{tab:pixel_utility}. Our method has high ($>90\%$) pixel utility comparing to other methods. Therefore, compared to baseline methods, GraphSeg retains a significantly larger number of pixels, as highlighted in \cref{fig:pix_util}. This means that GraphSeg can extract object level representations that are represented as much denser point clouds.

\begin{figure}[t]
  \centering
  \begin{subfigure}[t]{0.1\textwidth}
    \centering
    \includegraphics[width=0.98\textwidth]{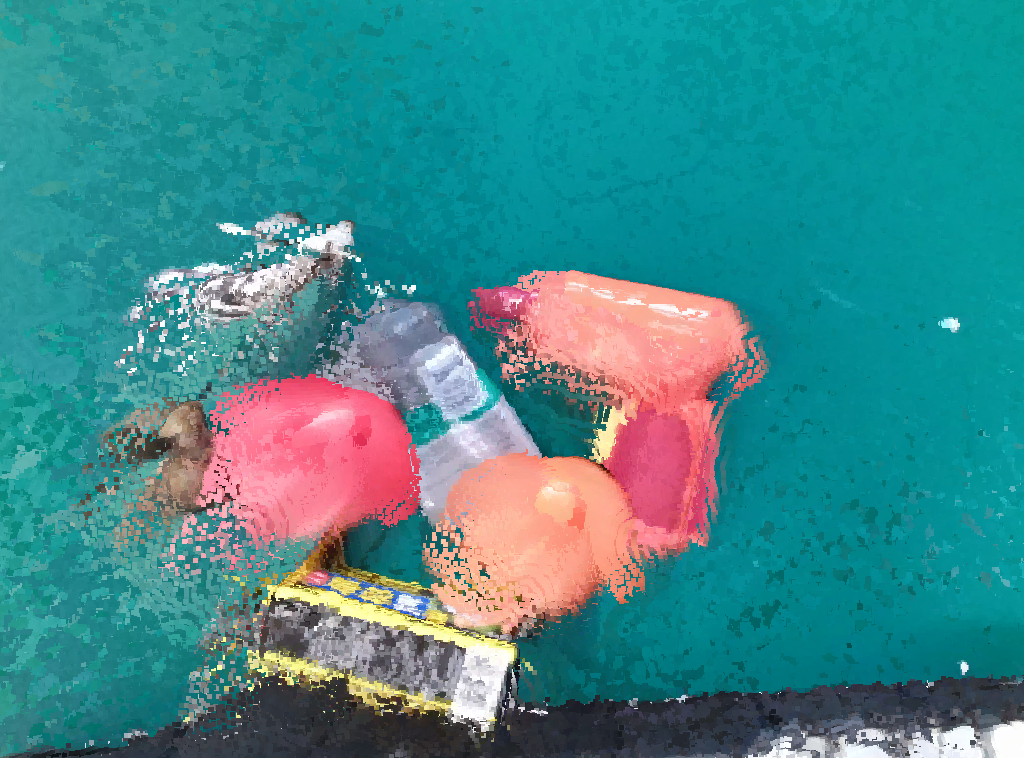}
    \caption{Input}
  \end{subfigure}%
  \begin{subfigure}[t]{0.1\textwidth}
    \centering
    \includegraphics[width=0.98\textwidth]{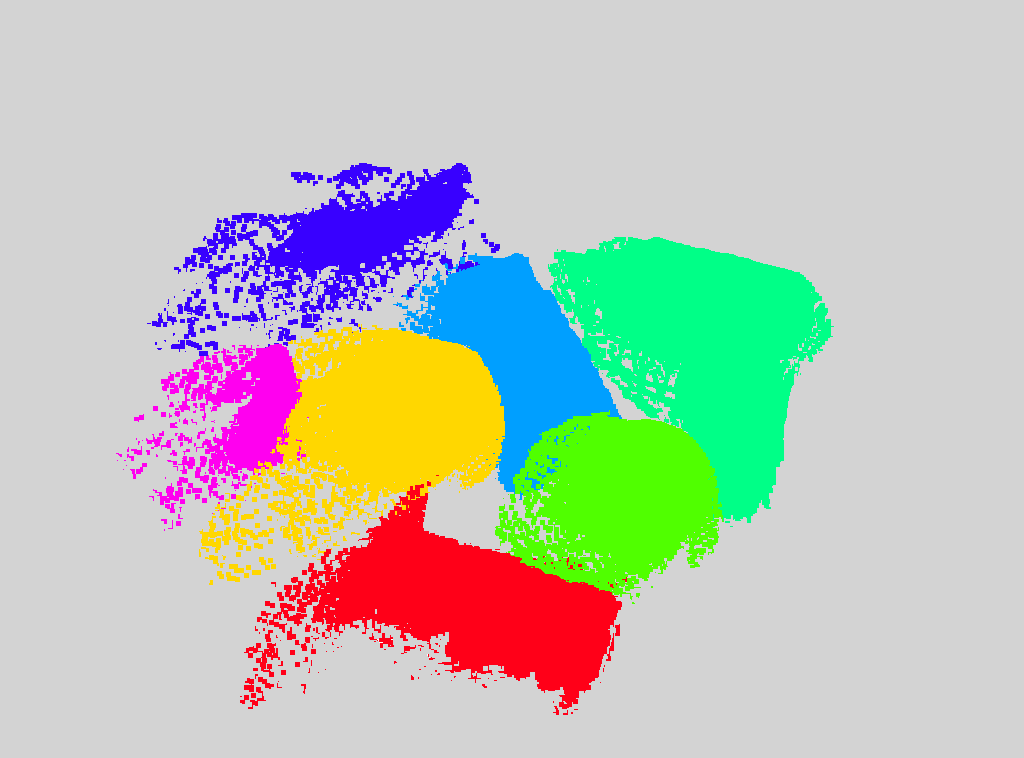}
    \caption{GT}
  \end{subfigure}%
  \begin{subfigure}[t]{0.1\textwidth}
    \centering
    \includegraphics[width=0.98\textwidth]{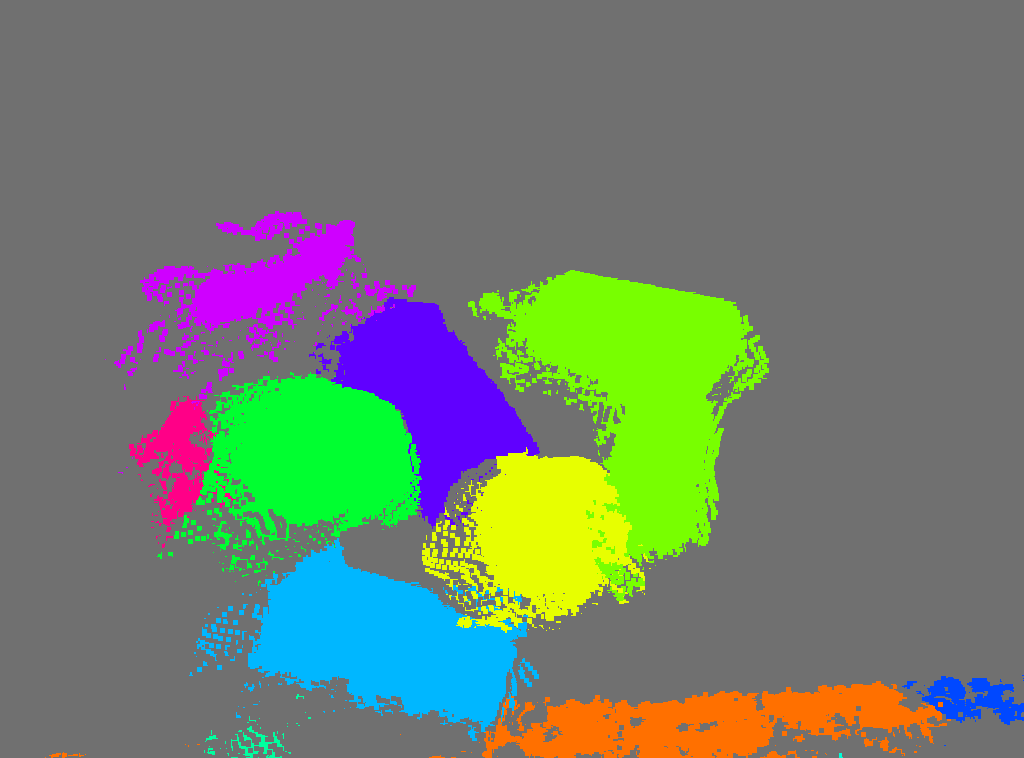}
    \caption{N=10}
  \end{subfigure}%
  \begin{subfigure}[t]{0.1\textwidth}
    \centering
    \includegraphics[width=0.98\textwidth]{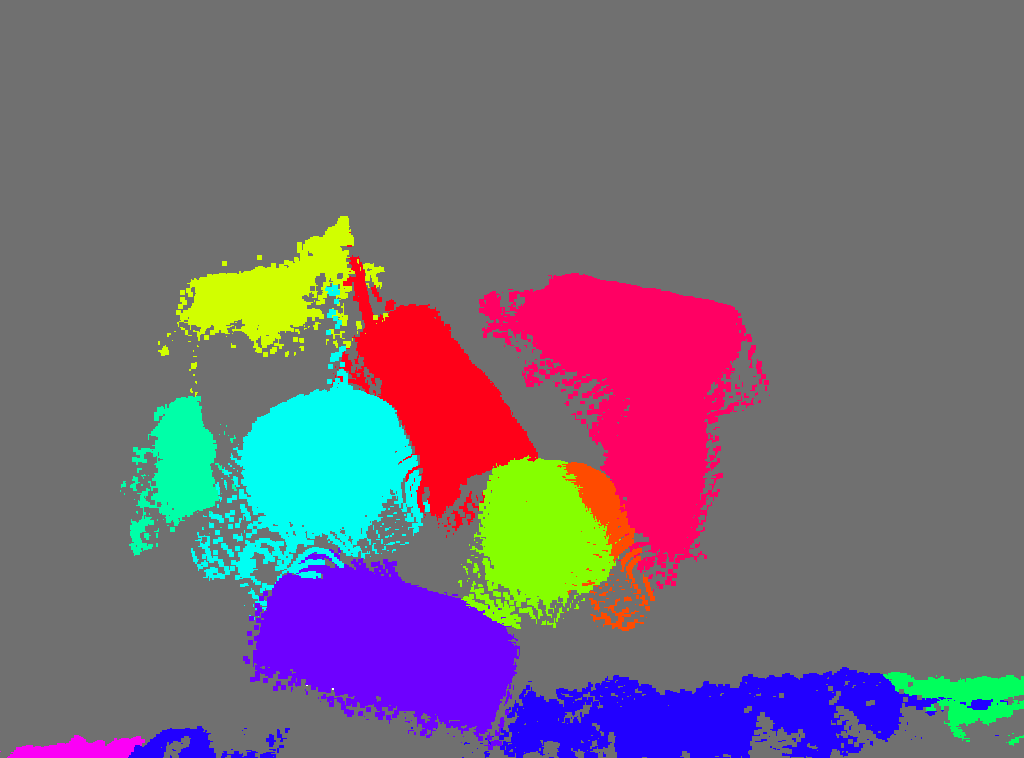}
    \caption{N=5}
  \end{subfigure}%
  \begin{subfigure}[t]{0.1\textwidth}
    \centering
    \includegraphics[width=0.98\textwidth]{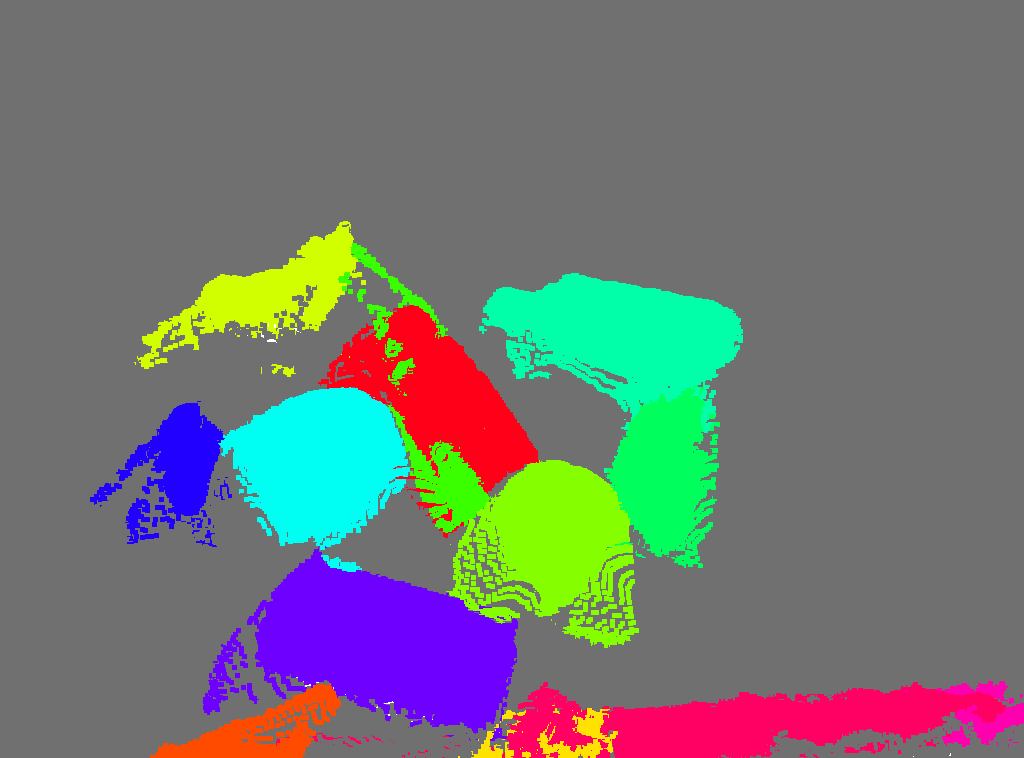}
    \caption{N=3}
  \end{subfigure}%

\caption{Sparse View Robustness: GraphSeg is also tested under sparse views. We observe robust performance even when the number of images (denoted by N) is small.}
\label{fig:sparser_view}
\end{figure}

\begin{figure}[t]
  \centering
  \begin{subfigure}[t]{0.2\textwidth}
    \centering
    \includegraphics[width=0.9\linewidth]{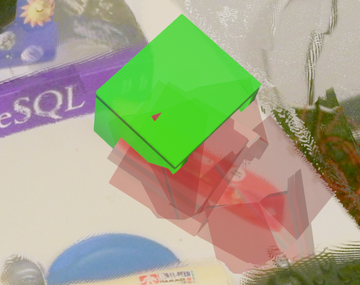}
  \end{subfigure}
  \begin{subfigure}[t]{0.2\textwidth}
    \centering
    \includegraphics[width=0.9\linewidth]{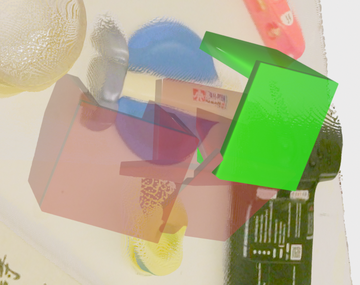}
  \end{subfigure}%
  \caption{Grasp pose candidates are generated using the grasp pose generator \cite{ten2017grasp} with the object's point cloud, then the candidates are filtered by the remaining point cloud by signed distance field collision-checking. The selected grasp is in green, and candidates in red.}\label{fig:grasp_filter}
\end{figure}

\begin{figure}[t]
  \centering
  \begin{subfigure}[t]{0.45\linewidth}
    \centering
    \includegraphics[width=0.48\linewidth]{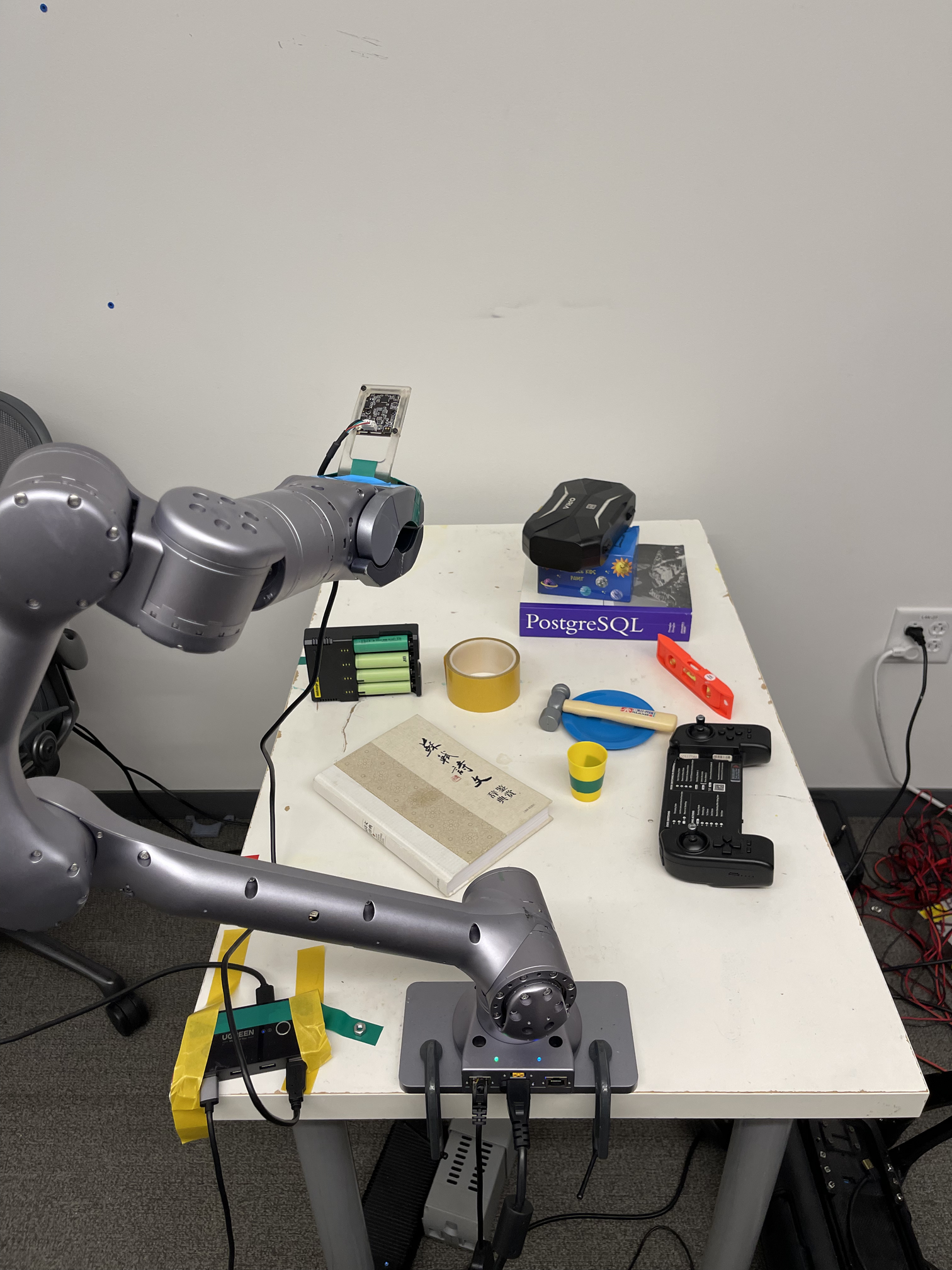}
    \includegraphics[width=0.48\linewidth]{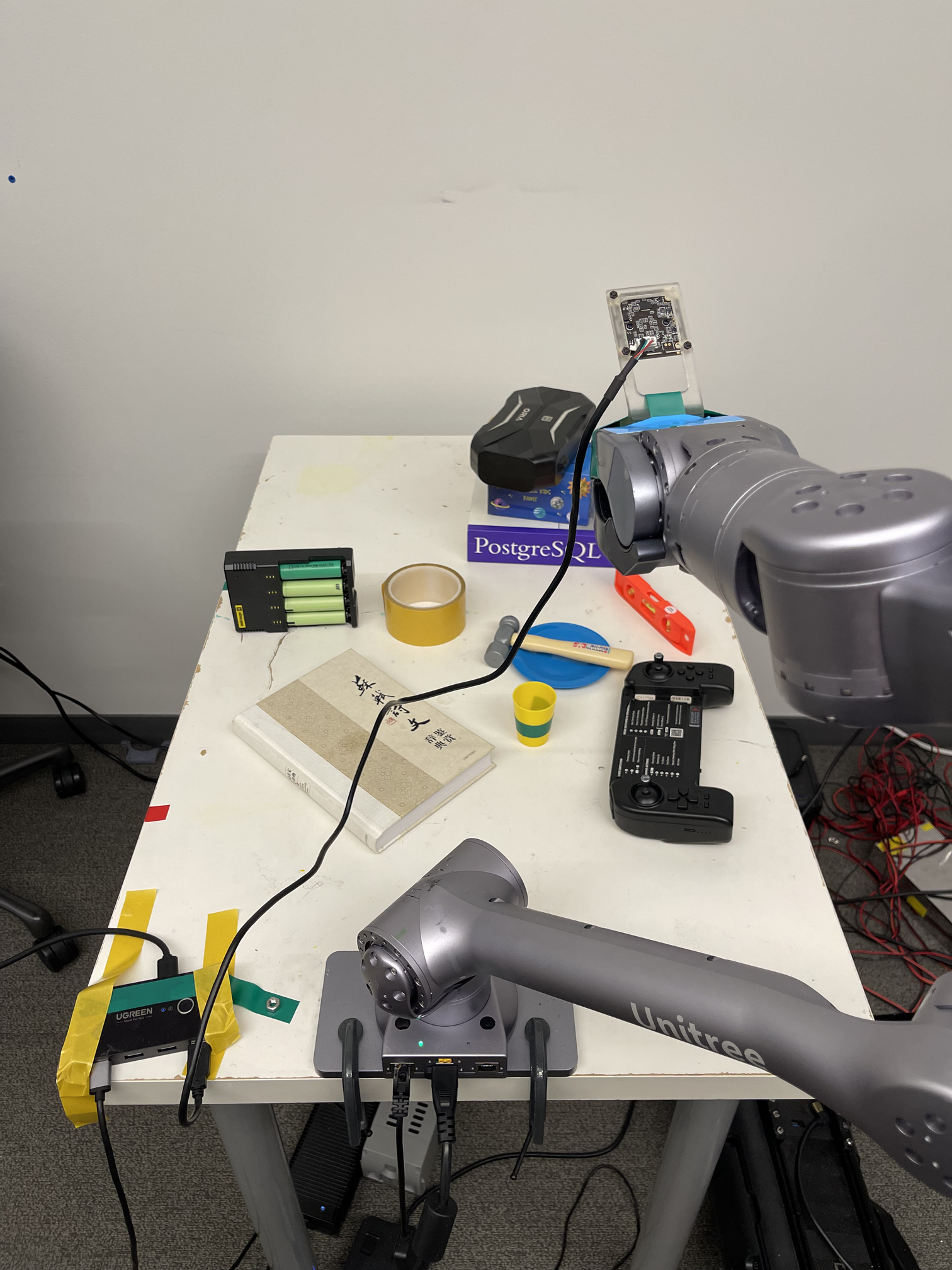}
    \caption{Setup}
  \end{subfigure}
  \begin{subfigure}[t]{0.38\linewidth}
    \centering
    \includegraphics[width=\linewidth]{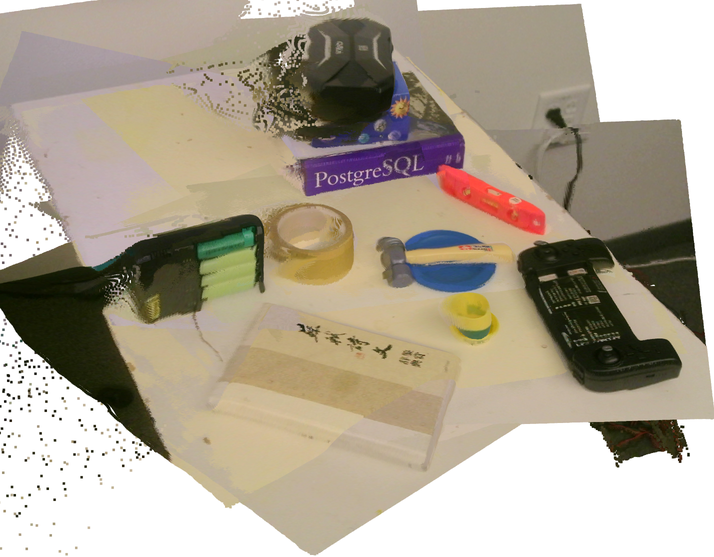}%
    \caption{3D Reconstruction}
  \end{subfigure}
  
  \begin{subfigure}[t]{0.85\linewidth}
    \centering
    \includegraphics[width=0.5\linewidth]{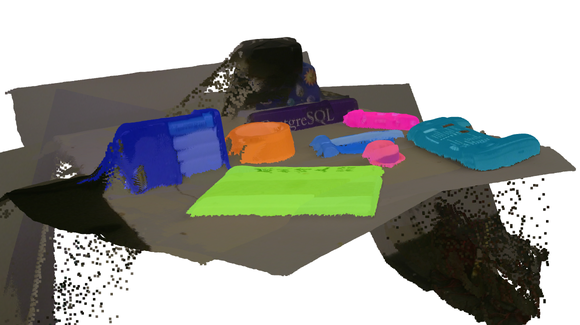}%
    \includegraphics[width=0.5\linewidth]{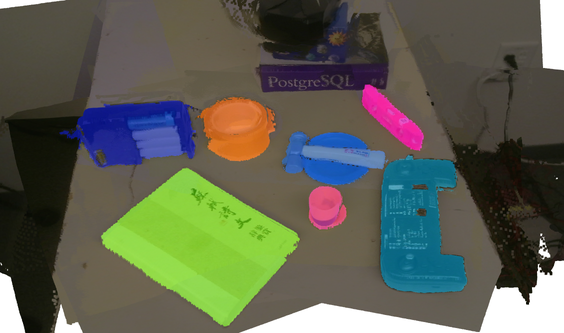}%
    \caption{GraphSeg Segmentation}
  \end{subfigure}
  \caption{(A) Real World Experiment Setup; (B) 3D reconstruction; (C) We run GraphSeg on the point cloud and RGB images, and output masked objects that we can grasp.}
  \label{fig:tabletop_setup}
\end{figure}

\subsection{GraphSeg is Robust Under Sparse-View}

To further evaluate the performance of GraphSeg, we test it under a more challenging scenario where the view is becoming increasingly sparse. Specifically, we evaluate the performance of GraphSeg, in which only $N=5$ and $N=3$ number of images are provided. This condition challenges GraphSeg with less pixel correlation and significantly less 3D spatial information. We visualize the result in \cref{fig:sparser_view}, from which we observe that GraphSeg retains its performance and provides reliable 3D segmentation even under sparse view conditions.

\subsection{Downstream Manipulation}
Segmentation of tabletop scenes to reveal object-level representations is crucial for robot manipulation. Typically, object representations, such as point clouds, are given to off-the-shelf grasp generators to generate feasible grasps for the robot. We demonstrate the effectiveness of GraphSeg on a real-world Unitree Z1 manipulator. A RGB camera is mounted on the end-effector of a robot manipulator, and objects are placed on the table top in front of the robot manipulator, then the robot manipulator moves around to take several RGB images of the tabletop. We utilize a 3D foundation model to generate point cloud and camera poses, and align it to the robot base via a hand-eye calibration method \cite{zhi2024unifying, hand-eye}; 

\begin{figure}[!t]
  \centering
  \begin{subfigure}[t]{0.1\textwidth}
    \centering
    \includegraphics[width=0.98\textwidth]{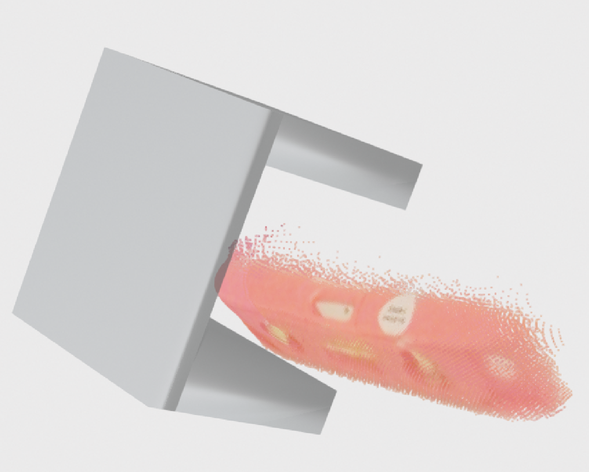}
    \includegraphics[width=0.98\textwidth]{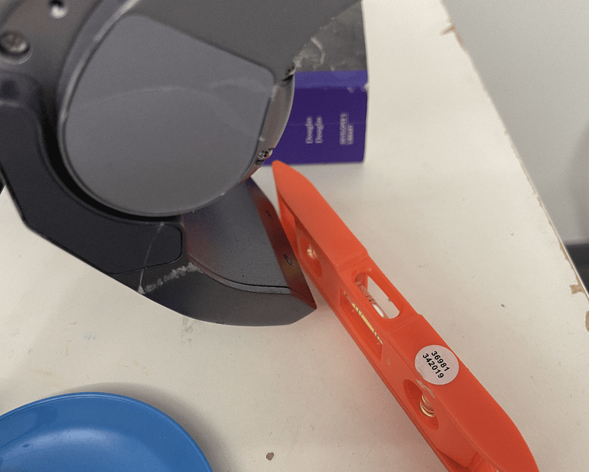}
    \includegraphics[width=0.98\textwidth]{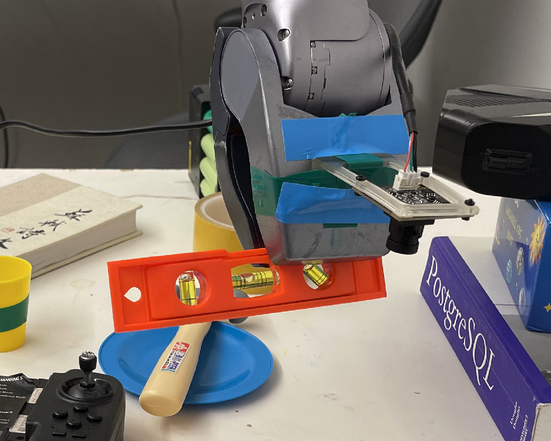}
  \end{subfigure}%
  \begin{subfigure}[t]{0.1\textwidth}
    \centering
    \includegraphics[width=0.98\textwidth]{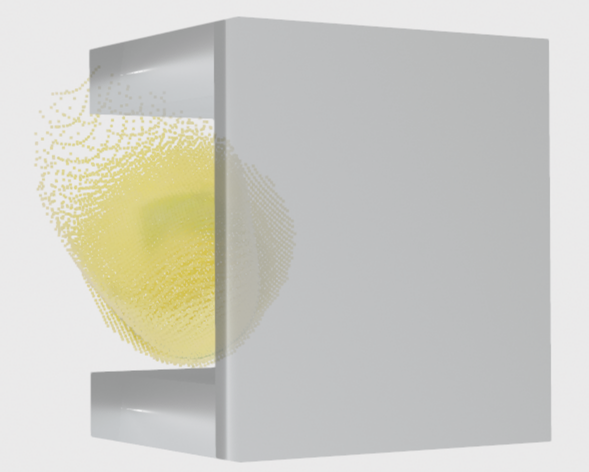}
    \includegraphics[width=0.98\textwidth]{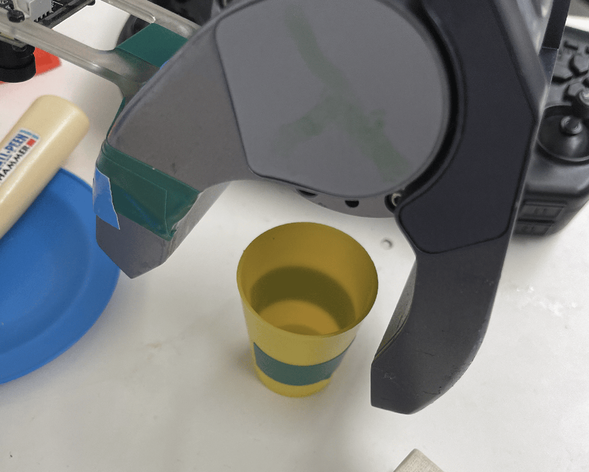}
    \includegraphics[width=0.98\textwidth]{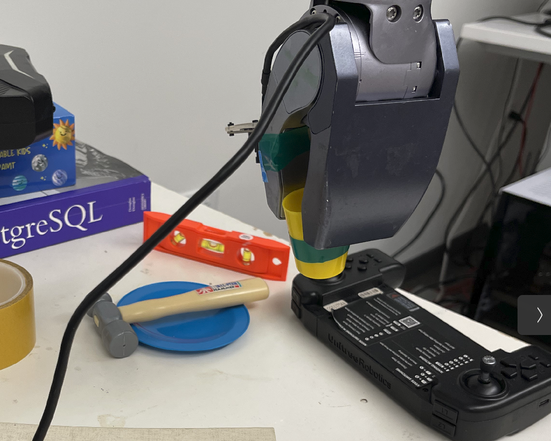}
  \end{subfigure}%
  \begin{subfigure}[t]{0.1\textwidth}
    \centering
    \includegraphics[width=0.98\textwidth]{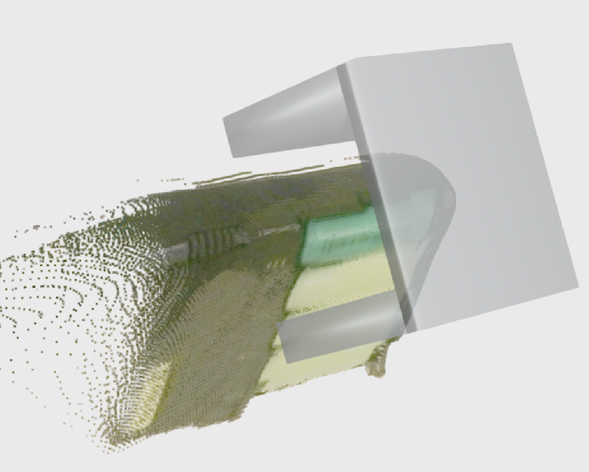}
    \includegraphics[width=0.98\textwidth]{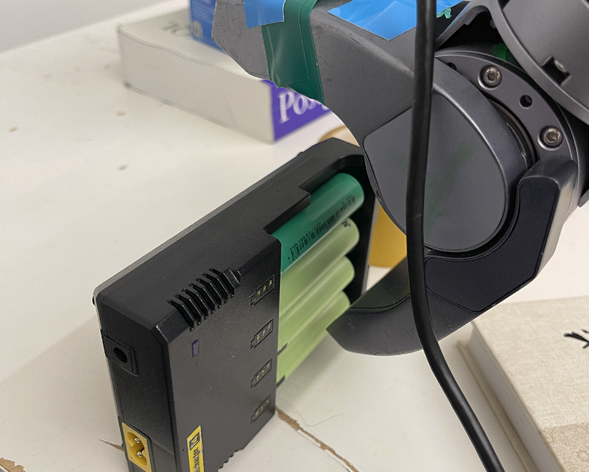}
    \includegraphics[width=0.98\textwidth]{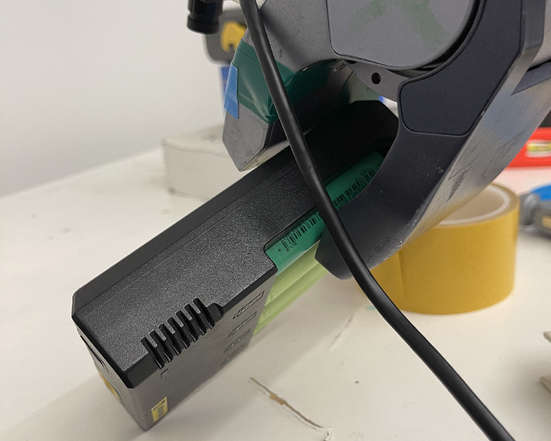}
  \end{subfigure}%
  \begin{subfigure}[t]{0.1\textwidth}
    \centering
   \includegraphics[width=0.98\textwidth]{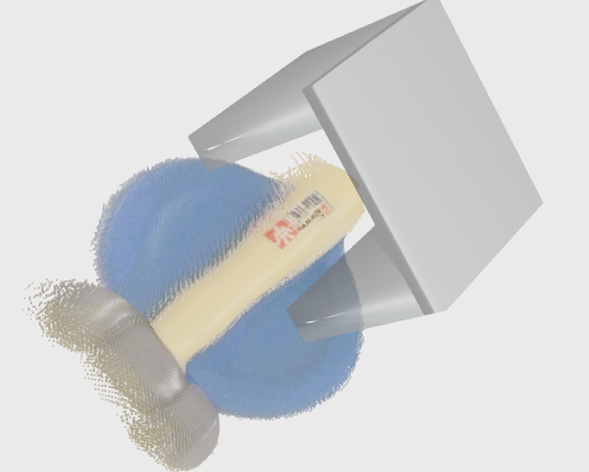}
    \includegraphics[width=0.98\textwidth]{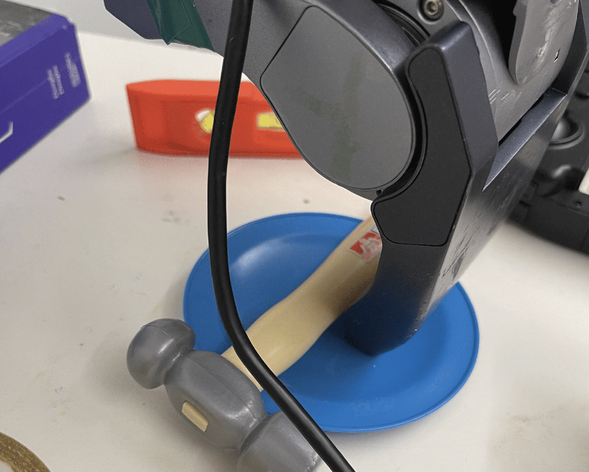}
    \includegraphics[width=0.98\textwidth]{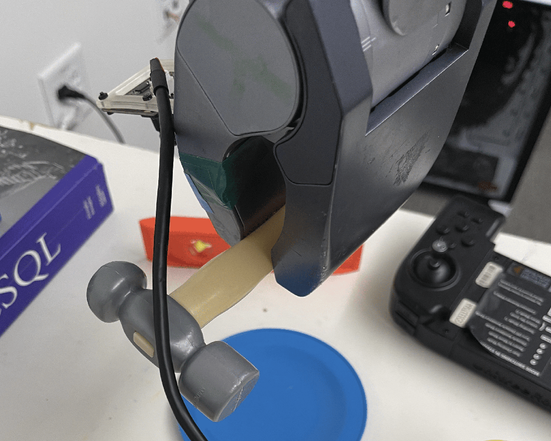}
  \end{subfigure}%
  \begin{subfigure}[t]{0.1\textwidth}
    \centering
    \includegraphics[width=0.98\textwidth]{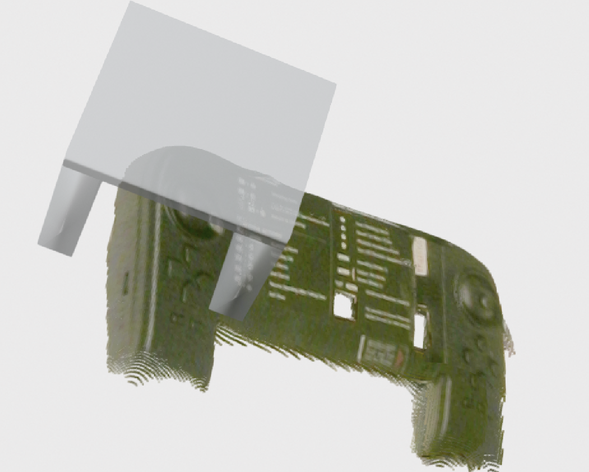}
    \includegraphics[width=0.98\textwidth]{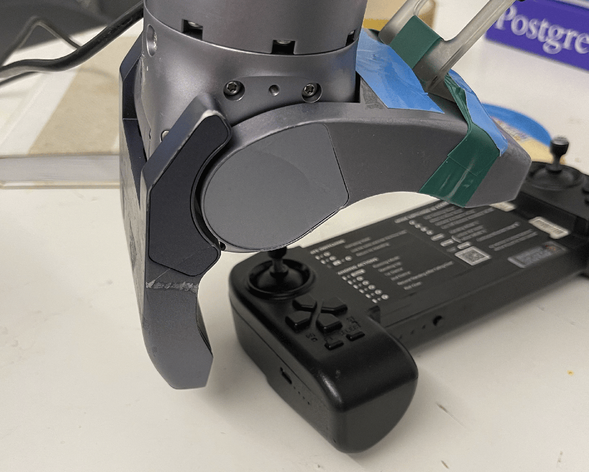}
    \includegraphics[width=0.98\textwidth]{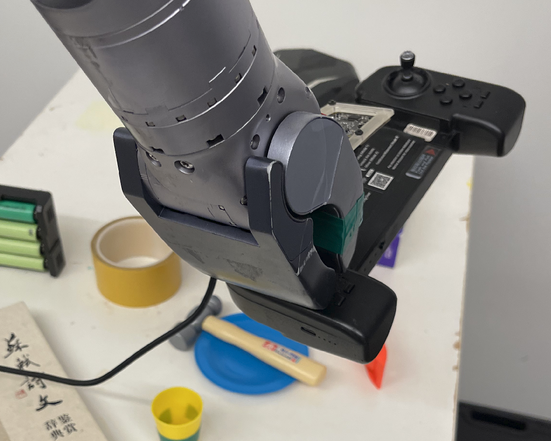}
  \end{subfigure}%
    \caption{Visualization of grasping various segmented instances. At the top, we illustrate the computed grasps, overlaid on the object-level representations built. We visualize the executed grasps for each object in the next two rows.}
    \label{fig: gras_viz}
\end{figure}

We run GraphSeg on the point cloud and RGB images, and output the segmentations of objects that we can grasp. We leverage GraphSeg to extract individual object point clouds from a set of multi-view RGB images of an unstructured table setup. These point clouds are then inputted, alongside gripper parameters, to a grasp generator. We use grasp generator \cite{ten2017grasp}, and filter infeasible solutions based on collision-checking against a signed distance field of the environment. An example is illustrated in \cref{fig:grasp_filter}. The robot then executes the corresponding grasps generated. We demonstrate successful grasp, on the real-world manipulator, for a variety of objects, including a surveyor's level, a hammer, a small cup, a battery package, and a remote controller. A planner can be integrated to generate collision-free motions \cite{Diff_templates, OMPL, diagrammaticlearning}. The experimental setup, 3D reconstruction, along with masks over nearby objects on the table, are all illustrated in \cref{fig:tabletop_setup}. In \cref{fig: gras_viz}, we illustrate the resulting grasps computed, over object-level representations, along with the successful grasps given in the subsequent rows.

\section{Conclusions and Future Work}
We propose GraphSeg, a framework that generates consistent 3D segmentations for tabletop scenes from a few RGB images. GraphSeg formulates segmentation as a graph edge–addition and contraction task, merging initially over-segmented 2D masks via both pixel-level and 3D structural correspondences. By leveraging 3D foundation models to recover scene geometry, GraphSeg preserves more object details and avoids over-segmentation. We empirically evaluate GraphSeg against state-of-the-art baselines, and demonstrate the performance and robustness of GraphSeg. We demonstrate that the dense object-level 3D representations produced enable downstream robot manipulation in the real-world. An avenue of future work can be to imbue uncertainty-aware behaviour, in a similar manner to probabilistic representations in \cite{HM,sptemp}, where the robot can actively take additional photos of regions that are ambiguous, so that the robot can improve the quality of the segmentation through placing the camera in more informative poses.

\bibliographystyle{ieeetr} 
\bibliography{bib}
\end{document}